\documentclass{ieeeaccess}
\usepackage{cite}
\usepackage{amsmath,amssymb,amsfonts}
\usepackage{algorithmic}
\usepackage{graphicx}
\usepackage{textcomp}
\usepackage{multicol}
\usepackage{subcaption}
\usepackage{color,soul}
\usepackage{hyperref}
\usepackage{soul}
\usepackage{color}
\usepackage{footmisc}
\usepackage[normalem]{ulem}
\makeatletter
\newcommand\footnoteref[1]{\protected@xdef\@thefnmark{\ref{#1}}\@footnotemark}
\makeatother

\def\BibTeX{{\rm B\kern-.05em{\sc i\kern-.025em b}\kern-.08em
    T\kern-.1667em\lower.7ex\hbox{E}\kern-.125emX}}
\begin{document}
%\history{ }
\doi{10.1109/ACCESS.2020.3014458}

\title{Machine Learning Based Analysis of\\Finnish World War II Photographers}
% \author{
% \IEEEauthorblockN{Kateryna Chumachenko\IEEEauthorrefmark{1}, Anssi M\"annist\"o\IEEEauthorrefmark{3}, Alexandros Iosifidis\IEEEauthorrefmark{2}, \textit{Senior Member, IEEE}, Jenni Raitoharju\IEEEauthorrefmark{1}, \textit{Member, IEEE}\\} 
% \IEEEauthorblockA{\IEEEauthorrefmark{1} Unit of Computing Sciences, Tampere University, Finland\\} 
% \IEEEauthorblockA{\IEEEauthorrefmark{3} Unit of Communication Sciences, Tampere University, Finland\\} 
% \IEEEauthorblockA{\IEEEauthorrefmark{2} Department of Engineering, Aarhus University, Denmark}}

\author{\uppercase{Kateryna Chumachenko}\authorrefmark{1},
\uppercase{Anssi M\"annist\"o}\authorrefmark{2}, \uppercase{Alexandros Iosifidis}\authorrefmark{3}, \IEEEmembership{Senior Member, IEEE}, and \uppercase{Jenni Raitoharju}\authorrefmark{1,4},
\IEEEmembership{Member, IEEE}}
\address[1]{Unit of Computing Sciences, Tampere University, 33014 Tampere, Finland}
\address[2]{Unit of Communication Sciences, Tampere University, 33014 Tampere, Finland}
\address[3]{Department of Engineering, 8000 Aarhus University, Denmark}
\address[4]{Programme for Environmental Information, Finnish Environment Institute, 40500 Jyväskylä, Finland}
%\tfootnote{This paragraph of the first footnote will contain support 
%information, including sponsor and financial support acknowledgment. For 
%example, ``This work was supported in part by the U.S. Department of 
%Commerce under Grant BS123456.''}

%\markboth
%{Author \headeretal: Preparation of Papers for IEEE TRANSACTIONS and JOURNALS}
%{Author \headeretal: Preparation of Papers for IEEE TRANSACTIONS and JOURNALS}

\corresp{Corresponding author: Kateryna Chumachenko (e-mail: kateryna.chumachenko@tuni.fi).}

\begin{abstract}
In this paper, we demonstrate the benefits of using state-of-the-art machine learning methods in the analysis of historical photo archives. Specifically, we analyze prominent Finnish World War II photographers, who have captured high numbers of photographs in the publicly available Finnish Wartime Photograph Archive, which contains 160,000 photographs from Finnish Winter, Continuation, and Lapland Wars captures in 1939-1945. We were able to find some special characteristics for different photographers in terms of their typical photo content and framing (e.g., close-ups vs. overall shots, number of people). Furthermore, we managed to train a neural network that can successfully recognize the photographer from some of the photos, which shows that such photos are indeed characteristic for certain photographers. We further analyzed the similarities and differences between the photographers using the features extracted from the photographer classifier network. We make our annotations and analysis pipeline publicly available, in an effort to introduce this new research problem to the machine learning and computer vision communities and facilitate future research in historical and societal studies over the photo archives. 
\end{abstract}

\begin{keywords}
Historical photo archives, Object detection, Photo framing, Photographer analysis, Photographer recognition 
\end{keywords}

\titlepgskip=-15pt

\maketitle

\section{Introduction}
\label{sec:introduction}
Historical photographs provide a valuable source of information for researchers in several fields of science. Alone the photographs of the two World Wars have been analyzed in archaeology \cite{WW_archaeology,WW_archaeologyII}, war history \cite{WW_ypres_salient,WW_warhistory2}, post‐phenomenological geography \cite{WW_repeat_photography}, photojournalism \cite{WW_Lapland,WW_photojournalism2}, religion \cite{WW_religion}, landscape research \cite{WW_landscape}, history of photography \cite{WW_photohistory,WW_history2}, propaganda research \cite{WW_propaganda,WW_propaganda2}, and others. Such research efforts require systematic analysis of large quantities of photographs, which is a laborious task taking a large part of the overall research time. State-of-the-art machine learning algorithms have potential to significantly speed up this task and also provide novel perspectives/directions for the following studies on different fields \cite{multimediaintelligence, cao2018recent, jiao2019survey}.

Despite the potential, up to this point the use of machine learning has been very scarce in this context. Previous works in the field include applications of face recognition to assist in identifying persons in historical portrait photographs \cite{hist_face_recognition}, feature matching for geolocalization or target matching in historical repeat  photography \cite{hist_match1,hist_match_dataset,hist_match2,hist_match3}, application of  marked point processes on automatic detection of bomb craters in aerial wartime images \cite{hist_craters}, a rudimentary classification of historical photographs into portraits, landscapes, group photographs, and buildings/architectural photography \cite{hist_classification}.

%Mohanty et al. \cite{hist_face_recognition} applied face recognition to assist in identifying persons in historical portrait photographs. Some studies \cite{hist_match1,hist_match_dataset,hist_match2,hist_match3} have applied feature matching for geolocalization or target matching in historical repeat photography. Kruse et al. \cite{hist_craters} applied marked point processes on automatic detection of bomb craters in aerial wartime images and Eiler et al. \cite{hist_classification} carried out a rudimentary classification of historical photographs into portraits, landscapes, group photographs, and buildings/architectural photography.

Wide-spread exploitation of machine learning in research using historical photographs has not started yet. One reason for this may be that the researchers performing such research typically have a background far from information technology. Besides not having the ability to use the novel machine learning tools, many researchers in these fields may not even realize the potential of machine learning in their work. Therefore, we demonstrate in this paper how state-of-the-art machine learning algorithms can assist and provide new insight in the historical photo analysis. As our case study, we concentrate on Finnish World War II photographs, while we use general algorithms and publicly available training data. Therefore, a similar analysis can be directly applied on any historical dataset.

The Finnish army produced a unique and internationally significant database of photographs during the Winter War, Continuation War, and Lapland War in 1939-1945. This collection is known as the Finnish Wartime Photograph Archive \cite{SAarchive} and it consists of almost 160,000 photographs captured by men who served in TK (Tiedotuskomppania = Information company) troops. The archive has been digitized in the beginning of 2010s and made publicly available in 2013. In its extent and historical significance, the Finnish Wartime Photograph Archive is comparable to the American Farm Security Administration/Office of War Information Photograph Collection \cite{FSAarchive}, which contains about 175,000 photos taken during the depression and drought in 1930s and World War~II. One of the official tasks of the TK troops was to collect ethnographic records. The Finnish Wartime Photograph Archive provides a unique cross section of the life especially in the Eastern Karelia occupied by Finnish troops during the Continuation War \cite{SarkynytArki}. 

The archive provides a valuable source of information for historians, photojournalists, and other researchers searching information of the life and sentiments behind the battles \cite{SAtutkimus}.  However, the original photograph labeling typically provides only the date, the place, the photographer, and a brief description of the key content. Thousands of photographs lack even this basic contextual information or it is incomplete. Moreover, not much of the content providing insight into the every day life and sentiments of the people has been originally described. Therefore, humanistic researchers have invested a considerable amount of time and effort to manually go through the collection and search for the information related to the studies at hand. In this paper, we show that machine learning algorithms can ease this kind of photo analysis, not only by helping to patch up gaps in the database but also by providing information that would be hard to obtain by manual inspection.

Several hundreds of photographers captured the Finnish Wartime collection. However, most of them only took one or few images and just a few dozen photographers captured half of the images. While the photographers did not have the freedom to select their topics freely, each photographer still provides a subjective view of the events. Objects appearing in the photos, scene setup, and picture framing vary based on professional background, personal training, and preferences of a photographer. Some of the photographers can be considered as skillful photojournalists or artists, while others simply recorded the events with their cameras with a less experienced approach.  Therefore, a better understanding of the differences of the individual TK photographers can provide deeper insight into the significance of the content and help researchers to find the content they are looking for. In this paper, we exploit state-of-the-art machine learning algorithms to analyze the characteristics and differences of 23 active TK photographers. We examine the typical objects appearing in the photographs and framing of the photos (i.e., close-ups vs. overall shots) for each photographer and we evaluate how distinguishable different photographers are.

In this work, our contribution lies on the edge of the historical photograph analysis and machine learning, allowing to make a step towards automatically answering historical research questions and to facilitate the work of historians. Rather than presenting a novel method from the machine learning perspective, we show how several common historical photograph analysis research problems can be addressed by utilizing modern machine learning techniques or making minor modifications to them. Our proposed approaches allow automating the work of historians that currently requires a large amount of monotonous manual labor. This can provide a significant speed up of the research process as well as the possibility to process considerably larger-scale data and allow historians to use their time on analyzing higher-level meaning and consequences of the gatherer results. More specifically,
\begin{itemize}
    \item we propose a pipeline for historical photograph analysis based on combination of four state-of-the-art object detection methods 
    \item we show how the above-mentioned pipeline can be utilized for photo framing evaluation
    \item we formulate a problem of photographer recognition and propose an approach for quantitative assessment of visual similarity of the photographers as well as establishment of unknown authorship  for photographs based on it  
    \item based on performed experiments, we provide the analysis of the most prominent Finnish WW2 photographers selected from Finnish Wartime Photograph Archive
    \item we provide the obtained bounding box annotations, codes, and all the pretrained models obtained in our study to facilitate further research in this area.\footref{data}
\end{itemize}
We have structured the rest of paper as follows: in Section~\ref{sec:mlhist}, we describe and discuss the tasks and methodologies adopted in this study in a general manner understandable also without previous knowledge on machine learning. We give the technical details separately in Section~\ref{sec:methods}, discuss the obtained results in Section~\ref{sec:results} and conclude the paper in Section~\ref{sec:conclusion}.

\section{Machine Learning for historical photograph analysis}
\label{sec:mlhist}
In this work, we propose and evaluate several application areas in which machine learning can assist in the analysis of historical images and photographers, namely, analysis of objects present in the scene, photo framing evaluation, photographer classification, and assessment of their visual similarity. The selected tasks illustrate only a small fraction of different ways machine learning can help in historical photograph analysis and were chosen based on their potential to provide a significant amount of useful information for researchers with a small amount of additional work that does not require deeper understanding of the underlying methods. We provide all the codes and models along with a detailed description on how to apply them on other historical photo archives\footnote{\label{data}We provide all codes, models, and obtained data annotations along with a detailed description on how to use them at \hyperlink{github.com/katerynaCh/Finnish-WW2-photographers-analysis}{github.com/katerynaCh/Finnish-WW2-photographers-analysis}. A permanent website will be created during the review process of this paper, which will host all information related to our research in this topic.}.

We selected for our experiments 23 Finnish war photographers. First 20 of them were the photographers with the highest total numbers of images in the Finnish Wartime Photograph Archive and the remaining three were included as they are considered by experts interesting for the photojournalistic research. The selected photographers along with the number of photographs and the photographing period for each photographer are listed in Table~\ref{tab:photographers}. The table also assigns photographer IDs used in later tables and illustrations. The total number of photographs considered in our analysis is $59,021$. It is likely that most of the photographers captured a higher number of photographs than suggested here. This is because thousands of photos in the Finnish Wartime Photograph Archive still lack the name of the photographer. As our analysis will help to differentiate the characteristics of the TK photographers, it may later contribute to suggesting names for at least some of the anonymous photographs.

\begin{table}[tb]
\caption{Selected photographers, total number of taken photographs, and photographing periods}
\begin{tabular}{l|llcc}
ID & Photographer & Total & Start date & End date \\
\hline
1      & Kim Borg            & 3932 & 25 Jun 1941 & 29 Oct 1944\\
2      & Tuovi Nousiainen    & 3551 & 25 Jun 1941 & 19 Sep 1944\\
3      & Ukko Ovaskainen       & 3523 & 24 Jun 1941 & 05 Jul 1944\\
4      & V\"ain\"o Hollming      & 3391 & 25 Sep 1941 & 09 Sep 1944\\
5      & Jarl Taube            & 3181 & 25 Aug 1941 & 11 Jul 1944\\
6      & Nils Helander         & 3125 & 14 Sep 1941 & 16 Jun 1944\\
7      & Pauli J\"anis            & 2903 & 10 Apr 1942 & 27 Sep 1944\\
8      & Oswald Hedenstr\"om       & 2812 & 24 Jun 1941 & 23 Sep 1944\\
9      & Esko Suomela          & 2755 & 25 Jun 1941 & 20 Sep 1944\\
10     & Tauno Norjavirta    & 2734 & 27 Jun 1941 & 21 Sep 1944\\
11     & Martin Persson          & 2615 & 02 Sep 1941 & 31 Aug 1943\\
12     & Kauko Kivi             & 2585 & 24 Jun 1941 & 02 Jul 1944\\
13     & Hugo Sundstr\"om        & 2564 & 24 Jun 1941 & 06 Nov 1944\\
14     & Vilho Uomala           & 2543 & 24 Jun 1941 & 20 Oct 1944\\
15     & Eino Nurmi            & 2379 & 25 Jun 1941 & 20 Aug 1944\\
16     & Holger Harrivirta       & 2307 & 26 Jun 1941 & 06 Dec 1942\\
17     & Olavi Aavikko          & 2109 & 10 Sep 1941 & 22 Jul 1944\\
18     & Uuno Laukka           & 1989 & 10 Aug 1941 & 10 Oct 1944\\
19     & Kalle Sj\"oblom          & 1967 & 20 Jun 1941 & 04 Aug 1944\\
20     & Pekka Kyytinen         & 1962 & 05 Jul 1941 & 15 Jul 1944\\
21     & Heikki Roivainen        & 1721 & 12 Sep 1941 & 21 Jul 1942\\
22     & Esko Manninen         & 1699 & 04 Jul 1941 & 20 Apr 1944\\
23     & Turo Kartto           & 674  & 17 Aug 1941 & 24 May 1942\\
\end{tabular}

\label{tab:photographers}
\end{table}
\subsection{Photo content analysis}
Presence of specific objects in a scene can provide plenty of information regarding an image of that scene. For example, it can be used for anomaly/fault detection \cite{staar2019anomaly, sarikan2018anomaly}, increasing autonomy of vehicles \cite{arnold2019survey}, and video surveillance \cite{huang2019deep}. To improve the performance of such scene analysis methods on new datasets, domain adaptation methods, that reduce the gap between the representation of the labeled training dataset and that of unlabeled target dataset, can be utilized \cite{hedegaard2020supervised, wang2020target, wang2019domain}. 

In the context of historical photo analysis, when using appropriate classes, detected objects allow to determine the context of each photo, as well as the focus of each photographer. For example, photographs on which chairs are detected, are likely to be taken indoors, while photographs on which horses, boats, cars, or trains are present, are more likely to be outdoor photos. Presence of objects such as skis can help determine the time of the year and, therefore, help establish the time period for unlabeled photographs. In turn, a high amount of chairs, ties, and people on the photo is likely to indicate a photo of some official event. At the same time, photos of airplanes are likely to resemble battles or near-battle areas. 

Besides analysis of a context of each specific photo, such analysis can help determine the main focus of each photographer by evaluating which types of objects are present in their photographs the most. For example, photographers having larger numbers of people, ties, and chairs, are more likely to have been urban photographers, while those with more animals (e.g., dogs, horses) are more likely to have worked in rural or countryside areas.

In our study, we propose to perform such analysis by leveraging the power of object detection methods, i.e., methods that are able to localize objects from a set of pre-specified classes on the images. Besides, we observe that labeling and training on a specific historical image dataset is often unnecessary as by combining the outputs of several strong object detectors pre-trained on public modern datasets one can achieve representative results for many common object classes. We also provide the obtained bounding box annotations for the Finnish Wartime Photograph Archive to facilitate further research in this area \footnoteref{data}.

\subsection{Photo framing evaluation}
The framing of a photograph is one of the stylistic decisions a photographer has to make. It is one of the most effective ways to assure visual variety in a group of photographs of a single situation. A traditional way of categorizing framings is to use three types as defined by Kobre \cite{Kobre}: overall shots, medium shots, and close-ups. A more detailed division of framings is widely used, e.g., in cinematic storytelling. According to this basic categorization, an overall shot sets the scene showing where the event took place: inside, outside, country, city, land, sea, day, night, and so on. This shot defines the relative position of the participants. A medium shot, on the other hand, should “tell the story” in one photograph by compressing important elements into one image. It is shot close enough to see the actions of the participants, yet far enough away to show their relationship to one another and to the environment. Finally, a close-up adds drama isolating one element and emphasizing it. In photographs of people, a close-up usually portraits a subject’s face. 

Measuring the ratio of different framings in a photographer’s works in a certain collection is one way to characterize his/her way of seeing. Here, we propose to take advantage of a combination of several object detectors for solving this task. To 
separate different framing categories, we examine the photographs with detected people and consider the relative size of the largest bounding box, which usually corresponds to the person closest to the camera, with respect to the image size.

\subsection{Photographer classification and visual similarity assessment}
Another problem in this study that we propose to address by utilization of machine learning techniques is the assessment of visual similarity of different photos and photographers. Ability to differentiate the photographers based on the visual cues of photographs can assist in labeling the images for which the author is unknown.
To evaluate how distinguishable different photographers are, we select a subset of 12 photographers (4-Hollming, 5-Taube, 6-Helander, 7-J\"anis, 8-Hedenstr\"om, 9-Suomela, 12-Kivi, 14-Uomala, 15-Nurmi, 19-Sj\"oblom, 21-Roivainen, 22-Manninen) and use some of the photographs from each photographer to train a neural network to recognize the photographer. 

A neural network sequentially applies a set of transformations to the input images, transforming them into such a representation that allows to distinguish different photographers, after which a classification layer classifies the image as belonging to a certain photographer. Therefore, besides directly establishing the authorship of unlabeled images, this approach allows to obtain such a feature representation of the photographs in which images that are visually similar are located closer in the feature space, allowing to asses the visual similarities of different photographers with each other as well as similarities of specific images quantitatively. 

For quantitative analysis, i.e., establishment of the extent to which the photographers are similar, we propose to utilize the Earth Mover's Distance \cite{emd} between the feature representations extracted from the pre-last layer of the neural network trained for photographer classification.

\section{Methods Description}
\label{sec:methods}
This section provides technical details on the machine learning approaches utilized for performing the previously described analysis, including object detection, photo framing analysis, photographer classification, and analysis of visual similarity of photographs.
\subsection{Photo content analysis and framing evaluation}
\label{ssec:aggregation}
For analysis of objects present in the scene of the photographs, we created a framework utilizing four state-of-the-art object detectors, namely Single-Shot Detector (SSD) \cite{SSD}, You Only Look Once v3 (YOLOv3) \cite{yolo}, RetinaNet \cite{retinanet}, and Mask R-CNN \cite{he2017mask}. We combine the four object detectors as such combination can result in improved detection accuracy and bounding box precision as compared to utilization of only one object detector, since generally a single object detector fails to detect all the objects of interest in the image. In this case, the information obtained from four independent detectors can compensate each other in terms of undetected objects of interest and therefore provide improved results. At the same time, combination of bounding box coordinates of objects detected by several object detectors can improve the precision of bounding boxes. All models were pretrained on MS-COCO dataset \cite{coco} that contains 80 classes. Among those, we considered people, airplanes, boats, trains, cars, bicycles, skis, dogs, horses, chairs, and ties as shown in Table~\ref{tab:detection}. The pipeline of aggregation of the information obtained from each object detector is described below in Sections III-A-1, 2, 3, and 4.

From each detector, we obtain a set of bounding boxes that are given as 4 coordinates and a class label with a corresponding confidence score. We discarded predictions with a confidence score below a certain threshold. This threshold was selected to be 0.7 for Mask R-CNN, 0.3 for RetinaNet, 0.5 for SSD, and 0.6 for YOLOv3. The thresholds were selected by manually investigating the effect of different scores in each detector on overall detection results. Higher threshold was selected for YOLOv3 and Mask R-CNN as they tend to produce more false positives with higher scores in our setup. 

In order to determine the final bounding boxes, the aggregation of the results from multiple detectors is performed. We investigate two combination approaches.  In each of the approaches, the identification of bounding boxes corresponding to the same objects is necessary. This is achieved as follows: first, the detected bounding boxes belonging to the same class are sorted according to their confidence scores. Then, the Jaccard similarities, also referred to as Intersection over Union \cite{iou} of each box with respect to the most confident bounding box of the class are calculated. 
Intersection over Union is defined by the area of the intersection of two boxes divided by the area of the union of these boxes:
\begin{equation}
    {IoU = \frac{Area\:\:of\:\:overlap}{Area\:\:of\:\:union}}
\end{equation}
The bounding boxes having the IoU more than certain threshold $\theta$ are identified as belonging to the same object. In our experiments, we set $\theta$ to 0.1. Then, the most confident bounding box and the boxes having IoU with it higher than the threshold are identified as belonging to the same object, marked as processed, and removed from the bounding box list, so that each bounding box is matched with at most one object. The process continues starting from the bounding box that has the highest confidence score after removal of boxes of the previous step until all bounding boxes have been processed.    

After this stage, we examined two options of combining the detections identified as belonging to the same object: either the bounding box with the highest confidence score can be selected, or the mean of each coordinate of all bounding boxes corresponding to the same object can be taken. Following the first approach, issues related to different scoring systems of different detectors arise, i.e., some detector might produce higher scores for all of its detections in general, while its bounding boxes might be less accurate. We also observe this heuristically, so in our experiments, we follow the second approach of taking the mean value of the coordinate produced by all the detectors and we observe that generally this results in more accurate positioning of the bounding box, although this cannot be evaluated quantitatively without the groundtruth information. This process was applied to bounding boxes of each class separately. 

We utilize the detections of the person class for photo framing evaluation, or, in other words, estimation of a distance from which the photo was taken. After combining the predictions of each detector, we used the largest bounding box for the person class in our photo framing evaluation. The evaluation is based on the area occupied by the bounding box - if the bounding box occupies more than 65\% of the overall photograph, the photo was classified as a close-up, 10-65\% - medium shot, and \textless 10\% - as an overall shot.

Further in this section, we provide details on the object detectors utilized
in the above-described approach.

\subsubsection{SSD}

The first object detector applied was SSD \cite{SSD} that is one of the most well-known single-shot detectors. The detector is based on the VGG-16 \cite{vgg} model pretrained on ImageNet dataset \cite{imagenet} that is used as a backbone feature extractor, followed by several convolutional layers that downsample the image and result in multiple feature maps. Using these feature maps from different layers, the detection can be done on multiple scales, while preserving the parameters across all scales, ensuring that both large and small objects are detected equally well. In addition to that, the single-shot approach results in high inference speed.

SSD relies on the idea of default bounding boxes, meaning that prior to training, several default bounding boxes are determined based on the amount of feature maps to be used and the size of the feature maps. Bounding boxes are created for the aspect ratios of $\{1,2,3,\frac{1}{2}, \frac{1}{3}\}$. During training, each groundtruth bounding box is associated with one of the default bounding boxes, determined by the highest Intersection over Union \cite{iou}. This default bounding box becomes a positive example for the groundtruth box, while the others become negative examples. 

At each scale, a feature map of different size is created and divided into a grid cell. During inference, a set of default bounding boxes is evaluated for each cell of the feature map and for each default bounding box, a shape offset is predicted along with the class probabilities for each class. Training is done with the combination of localization loss that is a Smooth L1 loss \cite{smoothl1} between the predicted box and the groundtruth box; and the confidence loss that is the cross-entropy loss over multiple class confidences. In our experiments, we used images rescaled to the size of $512\times512$ pixels as an input to SSD detector.

\subsubsection{YOLOv3}
The second object detector used was YOLOv3 \cite{yolo} that is in many ways similar to SSD: YOLO is a single-shot detector that makes predictions on multiple scales by performing detection on feature maps from different parts of the network. Prediction is done across three different scales obtained by dividing the image size by 32, 16, and 8.

YOLO relies on an ImageNet-pretrained Darknet-53 architecture that is used as a feature extractor backbone and multiple convolutional layers are added on top of it. Similarly to SSD, an image is divided into a grid cell and each cell is responsible for detecting the object, the center of which is located within its boundaries. Each grid cell predicts several bounding boxes along with the corresponding class label and confidence score. 

Rather than predicting bounding box coordinates directly, YOLO predicts the offsets from the predetermined set of boxes, referred to as anchors boxes or prior boxes, and each box is represented by the width and height dimensions \cite{yolov2}. These anchor boxes are obtained by applying $k$-means clustering \cite{kmeans} on the width and height dimensions of the boxes in the training set with the distance defined as
\begin{equation}
    d(box,centroid) = 1 - IoU(box, centroid),
\end{equation}
where both $box$ and $centroid$ are represented by two-dimensional vectors of width and height, $IoU$ stands for Intersection over Union, and $k = 9$ is chosen for $k$-means clustering, resulting in 9 anchor boxes. For calculation of $IoU$ we assume that the centers of the boxes are located at the same point. More specifically, for the model trained on COCO dataset and $416\times416$ images, the anchor boxes are $(10\times13),(16\times30),(33\times23),(30\times61),(62\times45),(59\times119),(116\times90),(156\times198)$, and $(373\times326)$.

For each detected bounding box, class prediction is obtained by multi-label classification with separate logistic classifiers. During training, the loss comprised of binary cross-entropy loss for object classification, and sum of squared error loss for bounding box prediction is used. YOLO operates on images of fixed size, and for our experiments all images were rescaled to $416\times416$ pixels size.

\subsubsection{RetinaNet}

The RetinaNet \cite{retinanet} object detector is the third state-of-the-art object detector used in this work. Overall architecture of RetinaNet consists of the backbone network for feature extraction, namely, Feature Pyramid Network \cite{fpn} built on top of ResNet \cite{resnet}, and two subnetworks, one of which is responsible for object classification, and the other one - for the bounding box regression. Similarly to previous detectors, the backbone network in pretrained on ImageNet dataset.

In a similar way to other detectors discussed so far, RetinaNet performs detection on multiple scales and relies on a predefined set of anchor boxes. Here, for each scale, anchors of 3 aspect ratios $\{1:2, 1:1, 2:1\}$ and 3 sizes $\{2^0, 2^{\frac{1}{3}}, 2^{\frac{2}{3}}\}$ are used, resulting in 9 anchor boxes per scale level. 

The subnet for object classification is a small fully-connected network, where the parameters are shared between different scale levels. The network is comprised of 3$\times$3 convolutional layers. For each spatial position, object class, and anchor box, a sigmoid activation function predicts the probability of presence of the object of that class. Thus, this subnet has the output of size $W\times H \times A*K$, where $A$ is the number of anchor boxes, $K$ is the number of classes, and $W$ and $H$ are the width and height of the corresponding feature map. The bounding box regression subnet is a fully-connected network that predicts four coordinates for each anchor box at each spatial location. The predicted coordinates correspond to the offset relative to the anchor. 

The main difference from other detectors lies in the utilization of the new loss function, referred to as Focal Loss, designed to address the issue of imbalanced classes in the object classification subnet:
\begin{equation}
FL(p_t) = -\alpha(1-p_t)^\gamma log(p_t);\:p_t = \begin{cases}p, \text{if } y=1\\
1-p, \text{otherwise} \end{cases}
\end{equation}
where $y = \pm 1$ is the ground-truth binary class label for the evaluated class, $p$ is the estimated class probability, $\gamma$ is a focusing parameter, and $\alpha$ is a balancing parameter. 
For the input to this detector, we rescaled the images preserving the aspect ratio and setting the size of the smaller side to 800 pixels, while keeping the size of a larger side at 1333 pixels maximum. 

\subsubsection{Mask R-CNN}
Mask R-CNN \cite{he2017mask} was the fourth detector used in this work. It is based on Faster R-CNN \cite{ren2015faster} - a region proposal based network consisting of two major blocks: a Region Proposal Network (RPN) that predicts the possible candidate locations of objects in the image, and a Region of Interest (RoI) classifier that extracts features of each candidate region proposed by RPN, assigns class labels to them, and refines the bounding box location.

Mask R-CNN extends Faster R-CNN for prediction of segmentation masks that is performed in parallel with bounding boxes prediction. Mask R-CNN predicts a binary segmentation mask for each candidate region proposed by RPN, resulting in $K$ of $m\times m$ masks per RoI, where $K$ is the number of classes. The prediction is achieved by Fully Convolutional Network. A per-pixel sigmoid is applied to the $m \times m$ mask output on the groundtruth class during training (i.e., only to the $c^{th}$ mask for the RoI with groundtruth class $c$), and the segmentation loss $L_{mask}$ is defined as an average binary cross-entropy loss. The total loss is defined as  $L = L_{cls} + L_{box} + L_{mask}$, where $L_{cls}$ and $L_{box}$ are the classification and bounding box regression loss, respectively, and they are defined in the same way as in original Fast R-CNN \cite{smoothl1}. 

Faster R-CNN relies on the RoIPool operations for extraction of small feature maps. RoIPool quantizes the float values of RoI into discrete bins to fit the granularity of the feature map, followed by spatial partitioning of the RoI into several spatial bins, to which pooling is applied. Such processing allows achieving higher training speed, while not affecting the performance much, as classification is robust to small translations. However, for the segmentation, pixel-accurate processing is required, resulting in the need for substitution of RoIPool with something else. For this purpose RoIAlign layer was proposed, where quantization is avoided: four locations are selected in each RoI bin and their values are computed using bilinear interpolation. Experimentally it is shown that usage of architecture with RoIAlign but without the mask segmentation component outperforms Faster R-CNN on bounding box prediction task already, and multi-task training for segmentation pushes the precision even further.

The architecture of Mask R-CNN consists of the convolutional backbone that is used for feature extraction, and a head that is used for classification, bounding box prediction, and segmentation.  In our setup, ResNet101 \cite{resnet} was used as a backbone, and FPN \cite{fpn} as the head. The image size of $540 \times 960$ was used for processing.

\subsection{Photographer recognition}

Evaluation of visual similarity of photographers and prediction of photograph authorship for unknown photos is achieved in this work by formulating an appropriate classification problem. For recognizing the photographer from the photos, we applied a pretrained and finetuned convolutional neural network. The architecture used in this work is a modified VGG-19 architecture \cite{vgg}, pretrained on ImageNet dataset. Modifications to the original architecture include the addition of Dropout layers after each pooling layer and each of the last two fully-connected layers with keeping 50\% of connections, and addition of a randomly-initialized fully-connected layer with 1024 neurons, followed by another Dropout layer that keeps 50\% of connections. At the final step, a layer with 12 neurons and softmax activation function is added. Adam optimizer was used for training with the learning rate of $10^{-5}$, momentum decay rates of $0.9$ and $0.999$ for the first and second moment estimates, respectively, and learning rate decay of $1e^{-6}$. 

\begin{figure*}
    \includegraphics[height=0.2\linewidth]{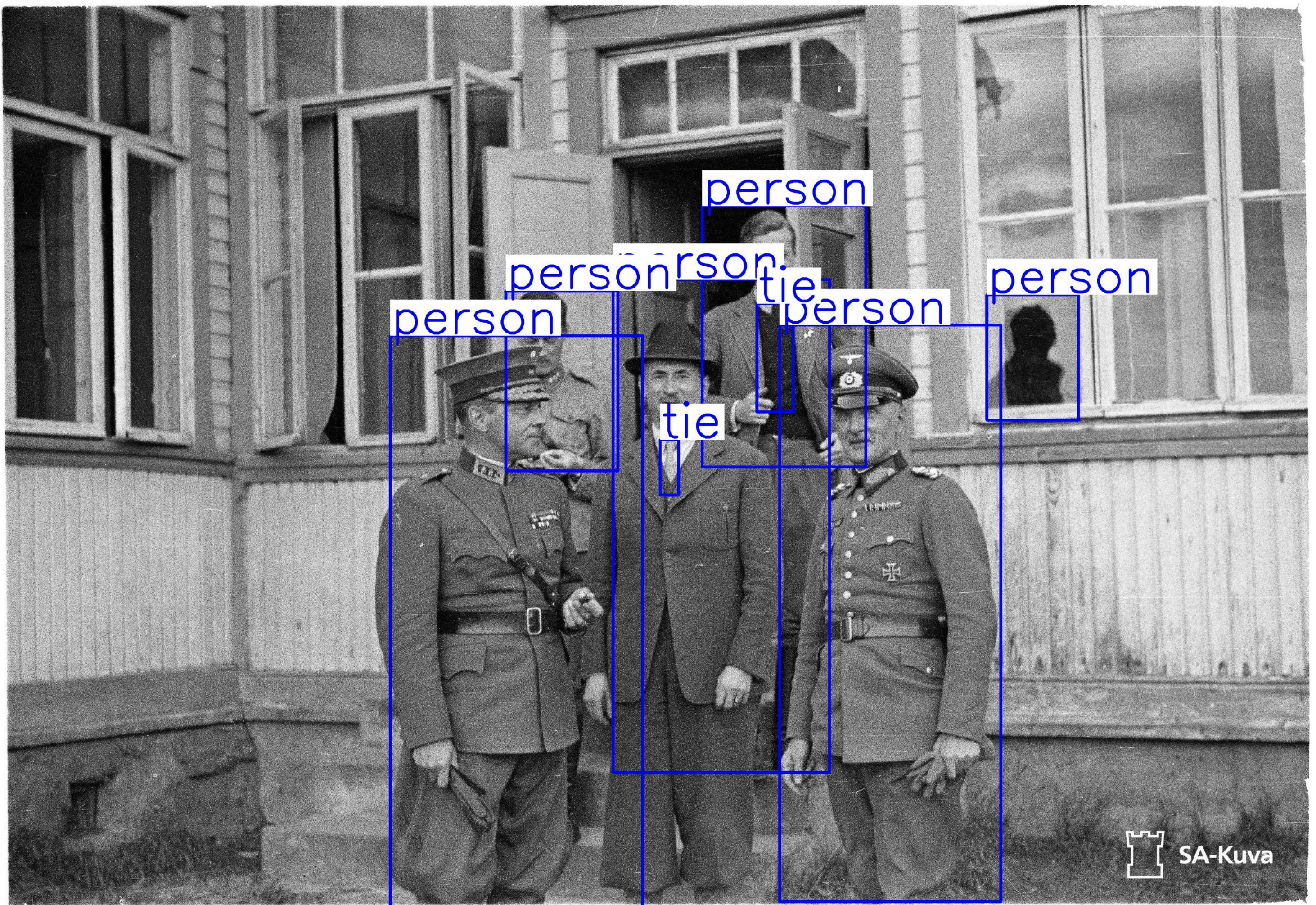}
    \includegraphics[height=0.2\linewidth]{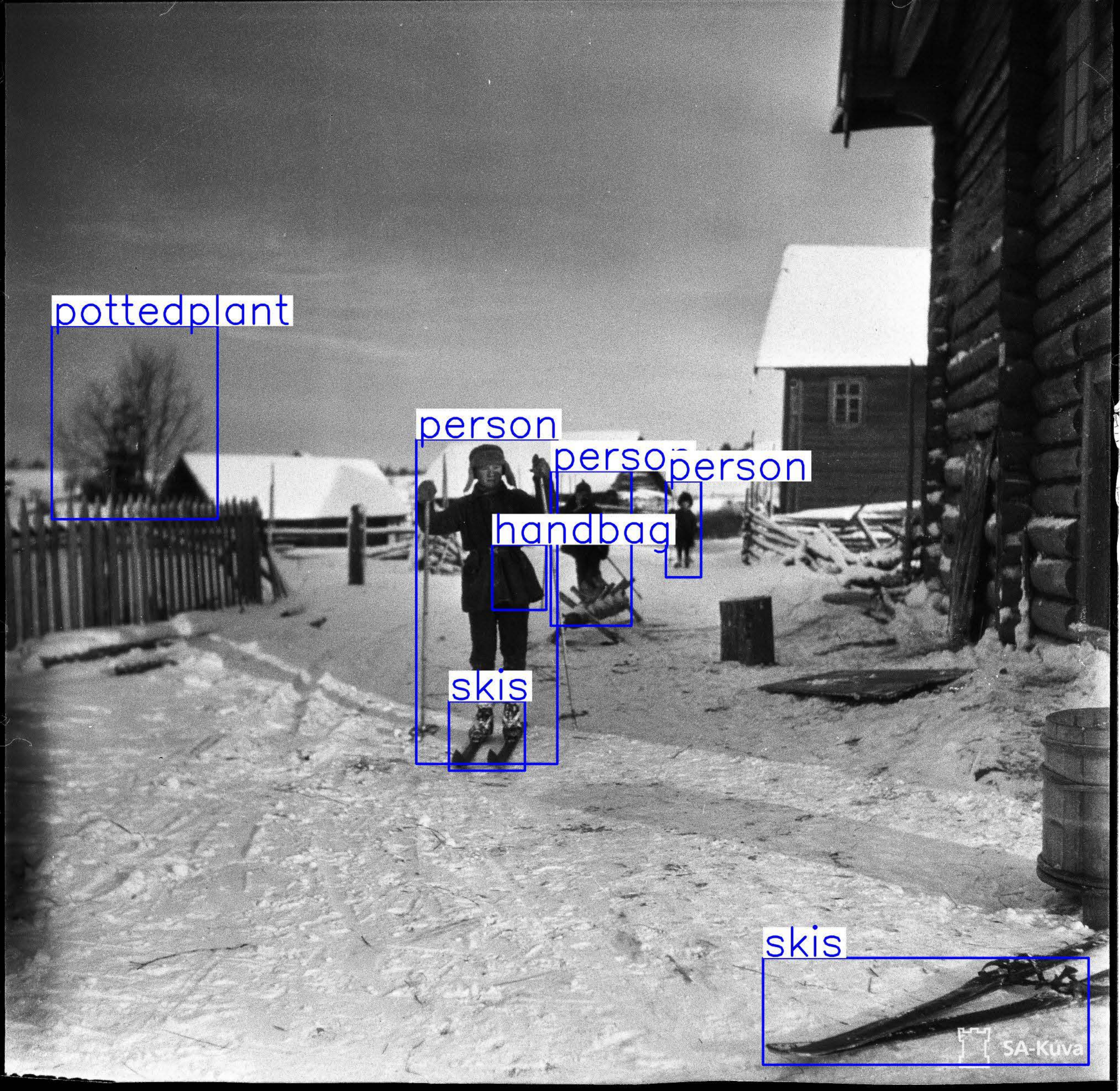}
    \includegraphics[height=0.2\linewidth]{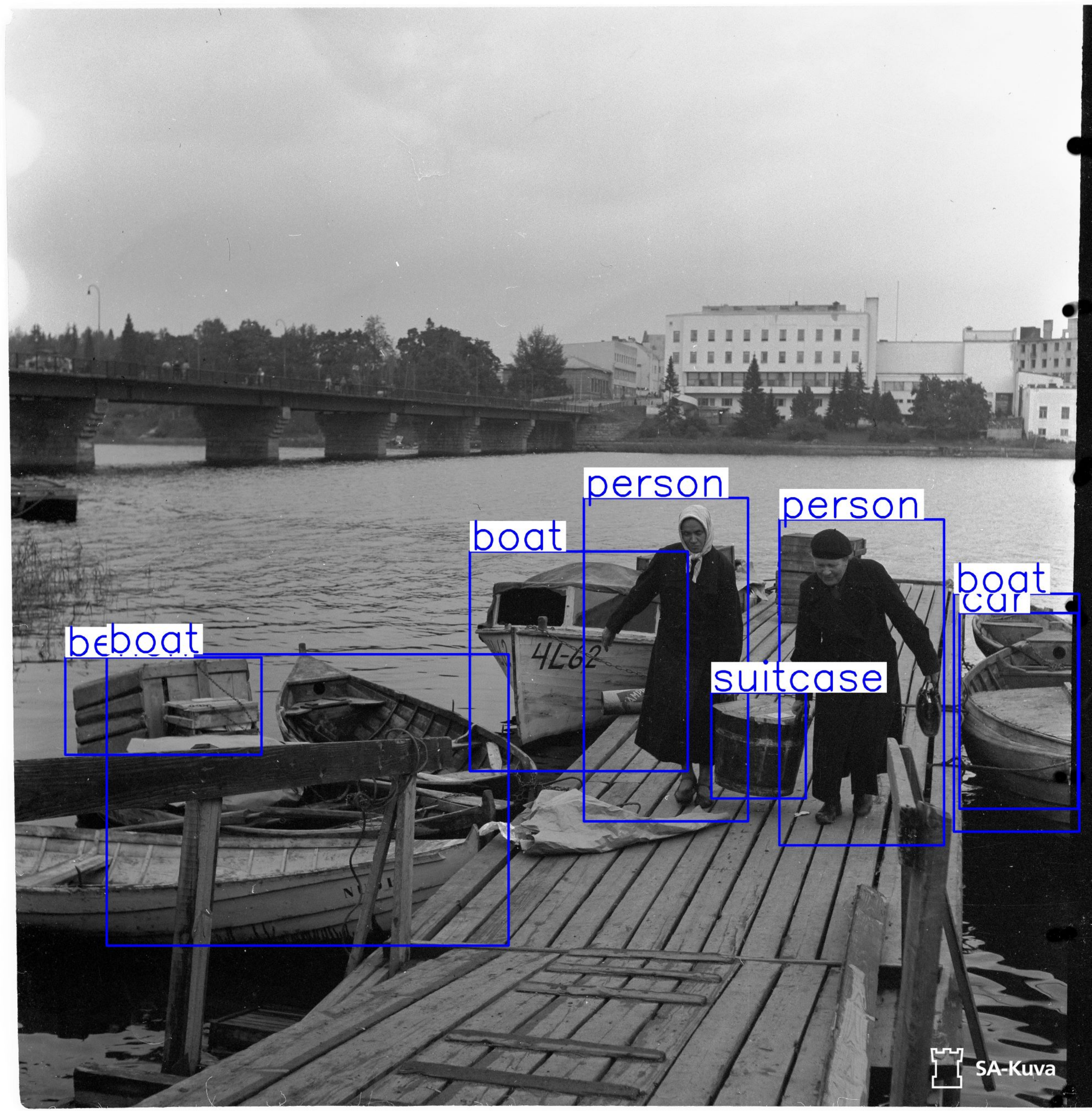}
    \includegraphics[height=0.2\linewidth]{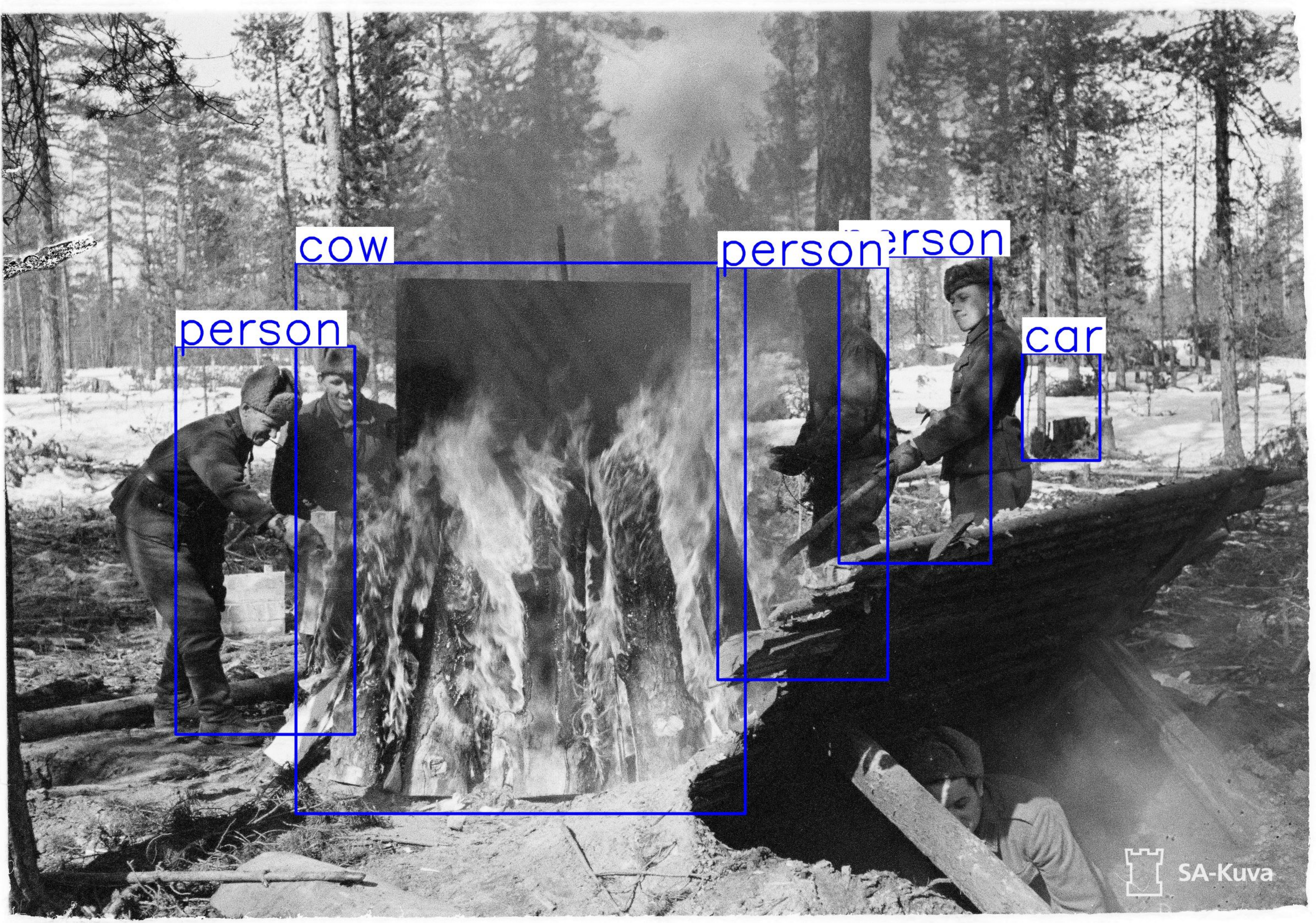}
    \caption{Examples of successful and erroneous object detection results. Histograms of the photographs shown here and in the following examples have been equalized. We show here also object classes not used in our analysis (e.g. cow).  Photographers: U. Ovaskainen, K. Borg, K. Borg, P. Jänis. Source of photographs: SA-kuva.}\label{fig:examples}
\end{figure*}

In order to address the issue of imbalanced classes, the weighted loss was used during training, calculated as: 
\begin{equation}
    \mathcal{L} = -w_c\frac{1}{N}\sum_{i=1}^N \:log(p[y_i \in C_{y_i}])
\end{equation}
\begin{equation}
    w_c = \frac{N}{N_c \times C},
\end{equation} where $N$ is the total number of training samples, $N_c$ is the number of training samples in class $c$,  $C$ is the total number of classes \cite{king2001logistic}, and $p([y_i \in C_{y_i}])$ denotes the predicted probability that $i^{th}$ observation $y_i$ belongs to the class $C_{y_i}$.

The training, validation, and test splits were selected randomly, while ensuring that the photos taken on the same day by the same photographer are not divided between splits, as they likely contain very similar photographs of a single event. In our setup, 60\% of the photos were selected as training set, 20\% - as validation set, and the rest - as the test set. As a preprocessing step, we performed histogram equalization on each photo on the value component in the HSV space in order to improve the contrast of each photo. Then, we resized the images into 224$\times$224 pixels size. Training was done for 100 epochs with batch size of 8 and categorical cross-entropy as the loss function.

\subsection{Photographer visual similarity}
In order to obtain a quantitative measure of visual similarity between photographers, we extract the features from the second last layer of the network trained for photographer recognition. Treating the set of features of each photographer as a signature of corresponding probability distribution, we calculate the Earth Mover's Distance \cite{emd, pyemd} between these distributions. The Earth Mover's Distance is defined as the minimal cost needed for transformation of one signature into the other, where the cost is based on some distance metric between two features. In our case, we utilize Euclidean distance. The Earth Mover's Distance between two distributions $P$ and $Q$ is then formally defined as:
\begin{equation}
    EMD(P,Q) = \frac{\sum^m_{i=1}{\sum^n_{j=1}f_{i,j}d_{i,j}}}{\sum^m_{i=1}{\sum^n_{j=1}f_{i,j}}},
\end{equation}
where $m$ and $n$ are the sizes of signatures of corresponding distributions, $f_{i,j}$ denotes the optimal flow between samples $i$ and $j$ found by solving the corresponding network flow problem, and $d_{i,j}$ denotes the distance between samples $i$ and $j$ \cite{emd}.

In order to visualize the relationships between the photos of different photographers, we utilize the same features that are used for calculating the photographer similarity. The resulting feature map has high dimensionality and for the visualization purposes we exploit the t-Stochastic Neighbour Embedding algorithm (t-SNE) \cite{tSNE}. t-SNE is a data visualization method for high-dimensional data, that aims at mapping the data instances in the high-dimensional space to some low-dimensional space, where the similarity between instances is preserved. This is achieved by modelling the similarities between instances as conditional probabilities. In the high-dimensional space, the similarity between data instances $x_i$ and $x_j$ is represented by the probability of $x_j$ to be selected as the nearest neighbor of $x_i$ if neighbors were selected proportionally to their probability density under a Gaussian distribution centered at $x_i$. In the low-dimensional space, instead of using the Gaussian distribution, the Student's t-distribution with one degree of freedom is used. Using a heavy-tailed distribution helps to model moderate distances in the high-dimensional space with a much larger distances in the low-dimensional space, resulting in better results compared to other methods. The Kullback-Leibler divergence of these probability distributions is then minimized with a gradient descent. The result of the visualization can be seen in Fig.~\ref{fig:tSNE}.

\section{Empirical study, Results, and Discussion}
In this section, we describe the experiments performed and discuss the obtained results for analysis based on object detection, photo framing evaluation, photographer classification, and their visual similarity assessment. 

\label{sec:results}
\begin{figure*}
    \centering
    \begin{subfigure}[b]{0.31\textwidth}
        \includegraphics[height=\textwidth]{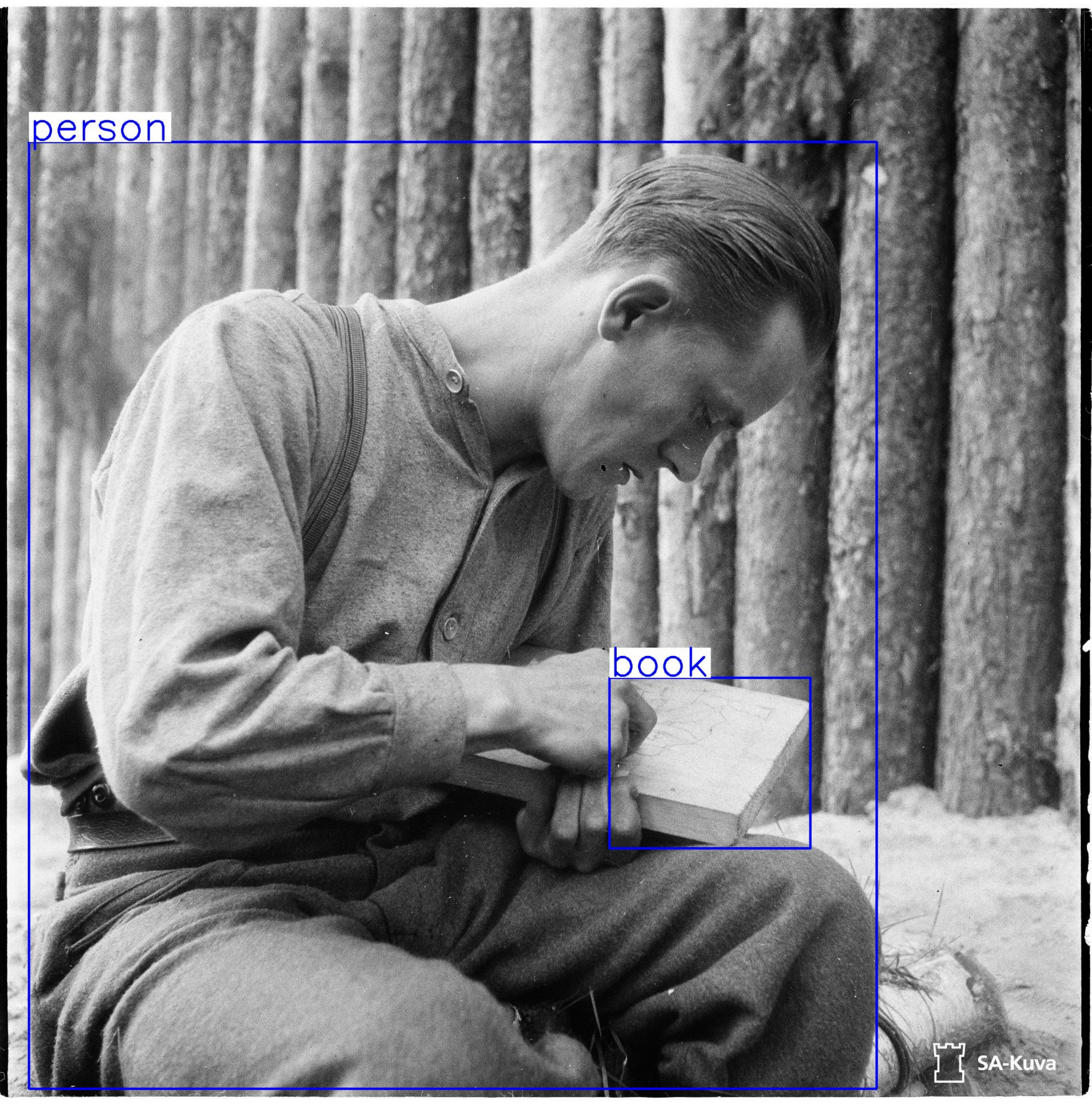}
        \caption{A close-up photo}
    \end{subfigure}
    ~ %add desired spacing between images, e. g. ~, \quad, \qquad, \hfill etc. 
    \begin{subfigure}[b]{0.31\textwidth}
        \includegraphics[height=\textwidth]{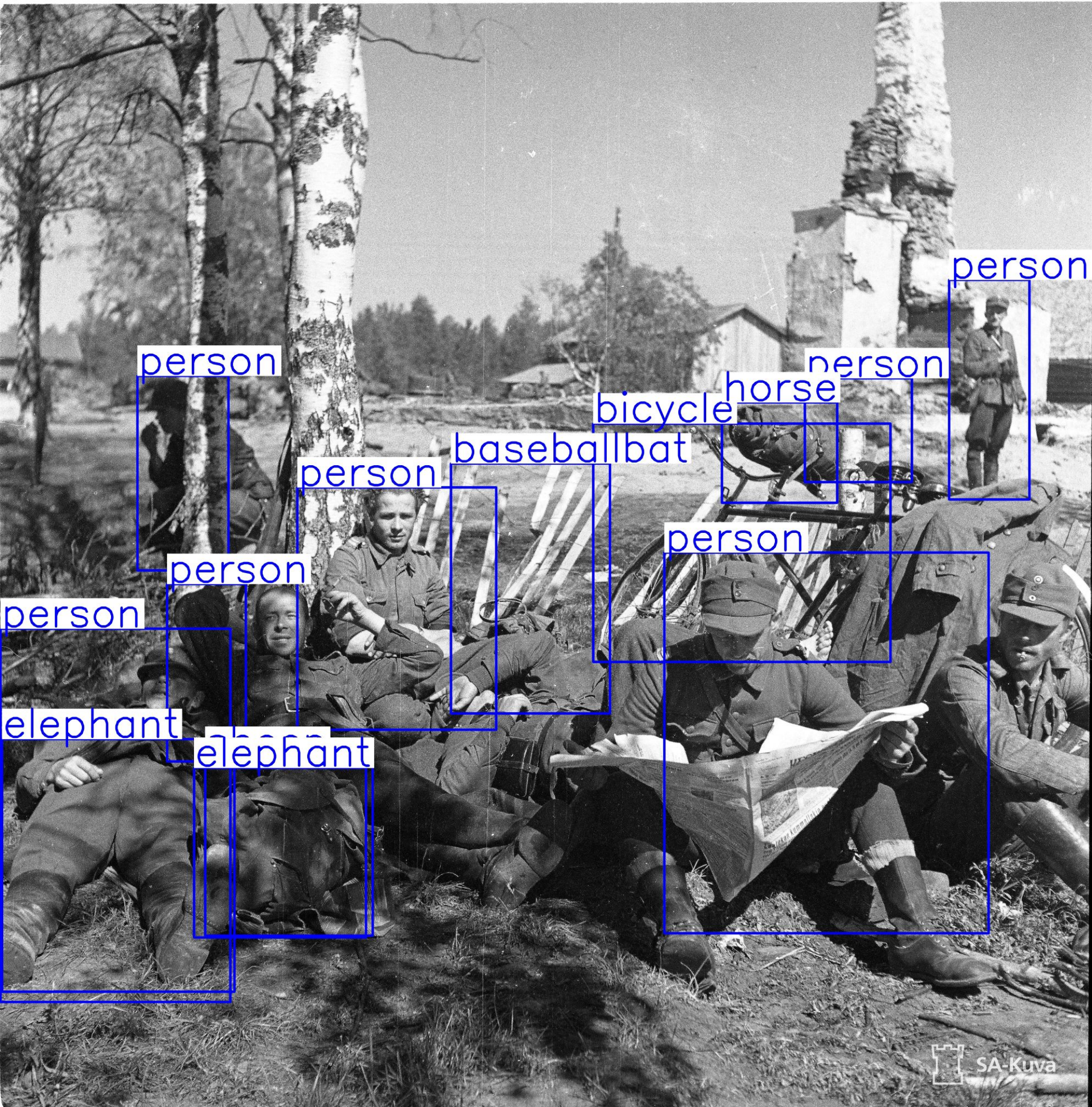}
        \caption{A medium shot}
    \end{subfigure}
    ~ %add desired spacing between images, e. g. ~, \quad, \qquad, \hfill etc. 
    \begin{subfigure}[b]{0.31\textwidth}
        \includegraphics[height=\textwidth]{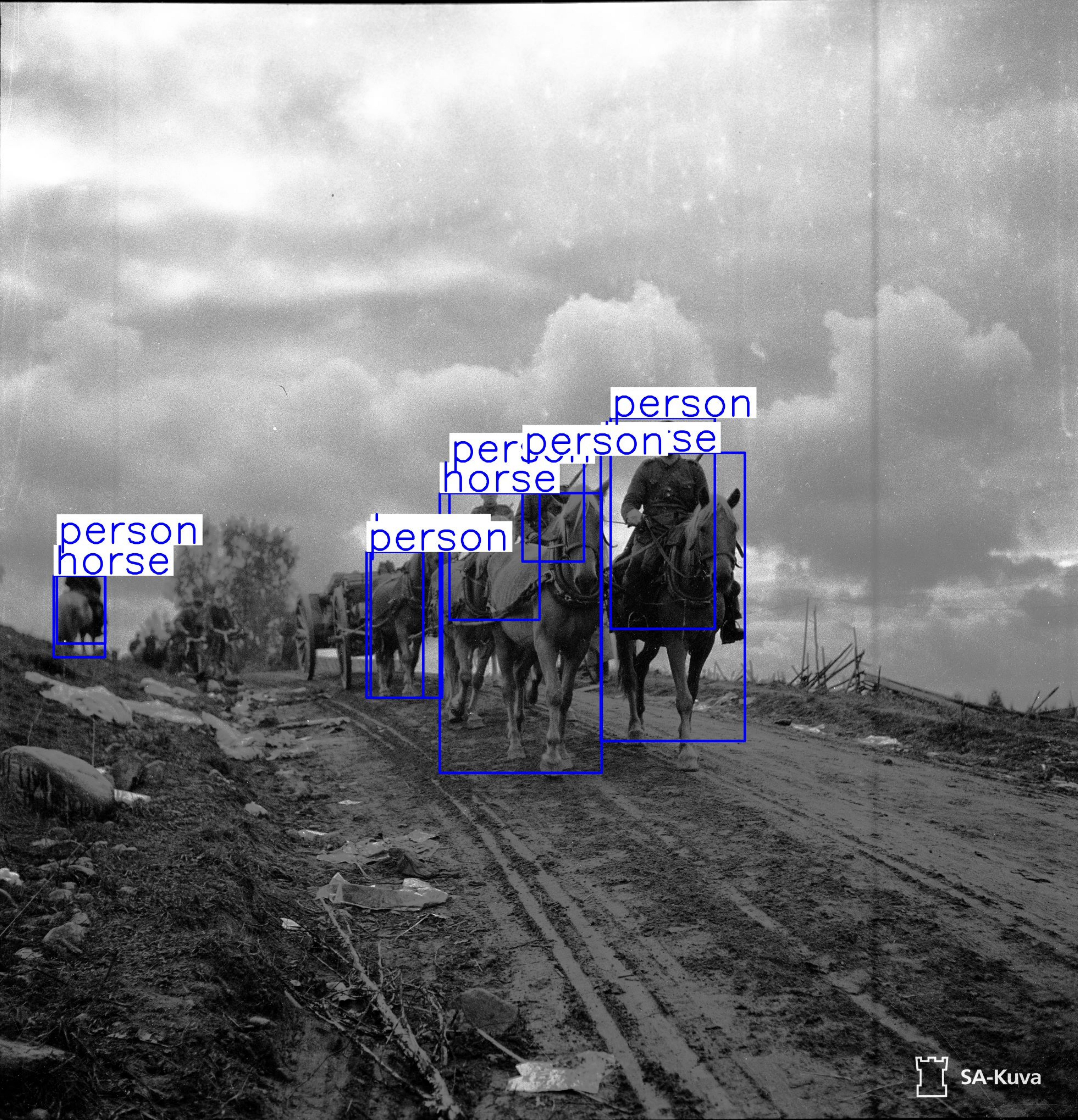}
        \caption{An overall shot}
    \end{subfigure}
    \caption{Examples photographs of different framing categories and the corresponding detection results. We show here also object classes not used in our analysis (e.g. elephant). Photographer: K. Borg. Source of photographs: SA-kuva.}\label{fig:ranges}
\end{figure*}
 
\begin{table*}[htbp]
\caption{Ratio of photos with people, number of people per such an images, and occurrences of other object classes per 100 images for different photographers}
\centering
\begin{tabular}{l|c|cc|cccccccccc|}
ID & Objects & Person & Persons & Airplanes & Boats & Trains & Cars & Bicycles & Skis & Dogs & Horses & Chair & Ties  \\
 & image & images & image & \multicolumn{10}{|c|}{100 images}  \\
\hline
1  & 6.8 & 0.89 & 2.8    & 2.5    & 9.4  & 6.3   & 8.7  & 6.2    & 2.4  & 4.6 & 12.1  & 11.9  & \emph{8.0}  \\
2  & 7.9 & 0.89 & 3.6    & 2.4    & \emph{4.1}  & \textbf{7.9}   & 9.8  & 7.2     & 1.8  & 3.1 & 8.1   & 24.8  & 14.5  \\
3 & 9.2 & 0.90 & 4.3    & 1.6    & 6.4  & 5.0   & \textbf{10.8}  & 7.1     & 2.5  & 5.7 & 18.1   & 16.6  & 22.5 \\
4  & 7.3 & 0.93 & 3.8    & 3.3    & 8.4  & \emph{1.6}   & 6.2  & \emph{2.9}     & \textbf{10.0}  & 6.7 & 15.7   & \emph{8.4}  & 13.0  \\
5  & 7.4 & \textbf{0.95} & 4.6    & 1.6    & 4.5 & 4.2   & 9.2  & 3.9  & 4.0  & 4.2 & 10.5   & 21.5  & 13.3 \\
6  & 7.8 & 0.90 & \emph{2.6}    & \textbf{14.5}    & 9.7  & 7.0   & 8.2  & 4.2     & 3.7  & 5.4  &  11.2   & 10.1  & 8.9  \\
7  & 6.8 & 0.91 & 3.7    & 2.4    & \emph{4.3}  & \emph{2.7}   & 6.6  & 3.4     & 3.5  & 5.7  & 16.1   & 14.0  & 12.6  \\
8  & 12.1 & 0.94 & \textbf{5.7}    & 3.1    & 8.8  & 5.8   & 8.7  & 4.8     & \textbf{6.5}  & \textbf{7.4} & 12.4   & 39.5  & \textbf{29.8}  \\
9  & 8.6 & 0.93 & 4.2    & 4.3    & \textbf{18.5}  & \textbf{7.5}   & 6.9  & 3.6     & 2.0  & 3.3 & \emph{5.2}   & 19.4  & 19.5  \\
10  & \emph{6.6} & \emph{0.85} & 3.6    & 2.0    & 8.4  & 4.3   & 9.5  & 3.8     & 1.3  & 4.0 & 13.4   & 9.0  & 13.1 \\
11  & 7.6 & 0.91 & 3.2    & 2.8    & 8.8  & 5.9   & 10.0  & 3.8     & 5.4  & 5.0 & 6.9  & 21.3 & 10.3  \\
12  & 9.8 & 0.92 & 4.0    & 7.4    & \textbf{15.7}  & 6.0   & 9.1  & \textbf{11.0}     & 6.2  & 7.0 & 19.1  & 12.8  & 9.6  \\
13  & 9.8 & 0.90 & 5.0    & \emph{1.5}    & 5.5  & 6.5   & 8.2  & 3.8     & 3.1  & \emph{3.0} & 7.9   & 36.4  & 27.0 \\
14  & 6.7 & \emph{0.84} & 3.8    & 2.9    & 11.6  & 4.8   & 6.4  & 4.0     & 4.5  & 5.4 & 13.6   & \emph{7.3}  & 9.6 \\
15  & 9.3 & 0.94 & 4.0    & \textbf{11.8}    & 7.6 & 4.1   & 8.1  & 4.3     & 1.8  & 5.3 & \textbf{27.0}   & 16.1  & 12.2 \\
16  & 6.7 & 0.89 & 3.2    & 3.1    & 9.7  & 4.8   & 10.2  & 3.4     & 2.1  & 5.1 & 16.7   & 8.7  & \emph{6.2}  \\
17  & 6.8 & 0.92 & 3.6    & 1.6    & 6.3  & 5.4   & 8.0  & 3.0     & 2.1  & 4.7 & 12.3   & 18.3  & 9.0 \\
18  & 7.2 & 0.90 & 3.8    & 1.7   & 7.3  & 3.7   & \emph{6.1}  & 4.8   & 4.5  & 5.5 & 9.9   & 16.8  & 14.5  \\
19  & 11.8 & \textbf{0.98} & 4.1    & 2.5    & 4.6  & 4.1   & 8.5  & 4.9     & 1.4  & \emph{1.8} & \emph{3.8}   & \textbf{53.7}  & \textbf{39.8} \\
20  & 7.3 & 0.88 & 3.5    & 1.7   & 12.5  & 4.2   & \emph{6.0}  & 3.7     & 1.3  & 4.5 & 7.6   & 17.1  & 17.7  \\
21  & 8.2 & 0.91 & \emph {2.7}    & 2.0    & 7.0  & 7.1 & \textbf{13.9}   & \emph{2.4}  & 5.6     & \textbf{8.5}  & \textbf{21.8} & 8.9   & 10.5   \\

22  & 14.8 & \textbf{0.95} & \textbf{6.1}    & 3.4   & 11.7  & 5.3   & 6.8  & \textbf{8.0}     & 1.3 & 4.4  & 10.0   & \textbf{79.1}  & 27.0  \\

23  & 8.1 & 0.94 & 3.1    & \emph{1.5}    & 5.2  & 6.4   & 8.2  & 7.3     & \emph{1.0}  & 5.0 & 17.8   & 15.5  & 17.5 \\
\hline
Avg & 8.5 & 0.91 & 3.3 & 3.6 & 8.5& 5.2 & 8.4 & 4.9 & 3.4& 5.0 & 12.9 & 21.2 & 17.5
\end{tabular}

\label{tab:detection}
\end{table*}

\subsection{Photo content analysis}

We applied pretrained object detection algorithms to detect the objects appearing in images. Out of the available 80 object classes, we manually selected 11 relevant classes (people, airplanes, boats, trains, cars, bicycles, skis, dogs, horses, chairs, and ties). We also empirically checked that the detection quality for these classes was high. Some of the potentially interesting classes, e.g., cow, we discarded, because many cow detections were actually horses, reindeer, or other objects. Also for the selected classes, the results should be considered only as indicative. When objects are clearly visible, they are typically well detected. However, there are cases where objects are missed or misidentified. Few examples of object detections are shown in Fig.~\ref{fig:examples}.
\begin{figure*}
    \centering
    \begin{subfigure}[b]{0.45\textwidth}
        \includegraphics[width=\textwidth]{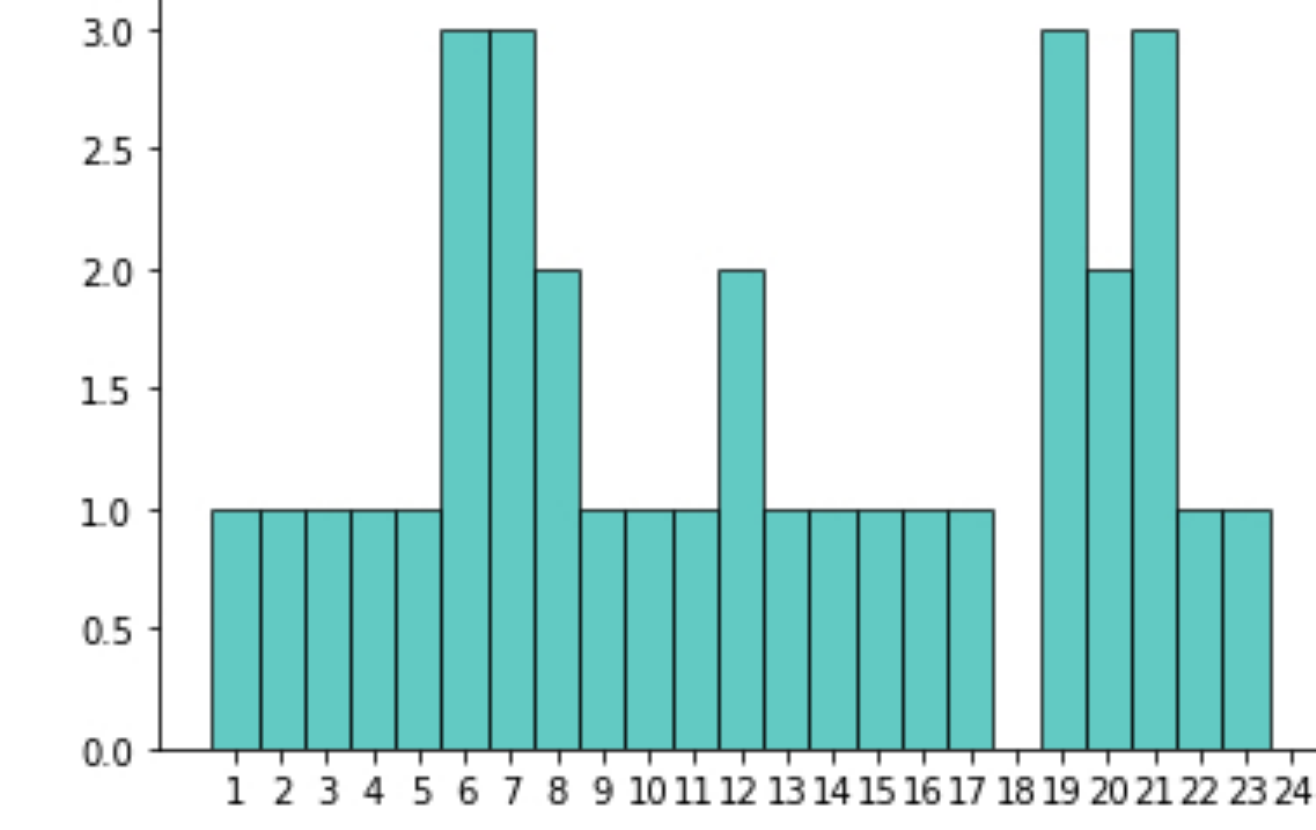}
        \caption{Percentage of close-ups}
    \end{subfigure}
    ~ %add desired spacing between images, e. g. ~, \quad, \qquad, \hfill etc. 
    \begin{subfigure}[b]{0.45\textwidth}
        \includegraphics[width=\textwidth]{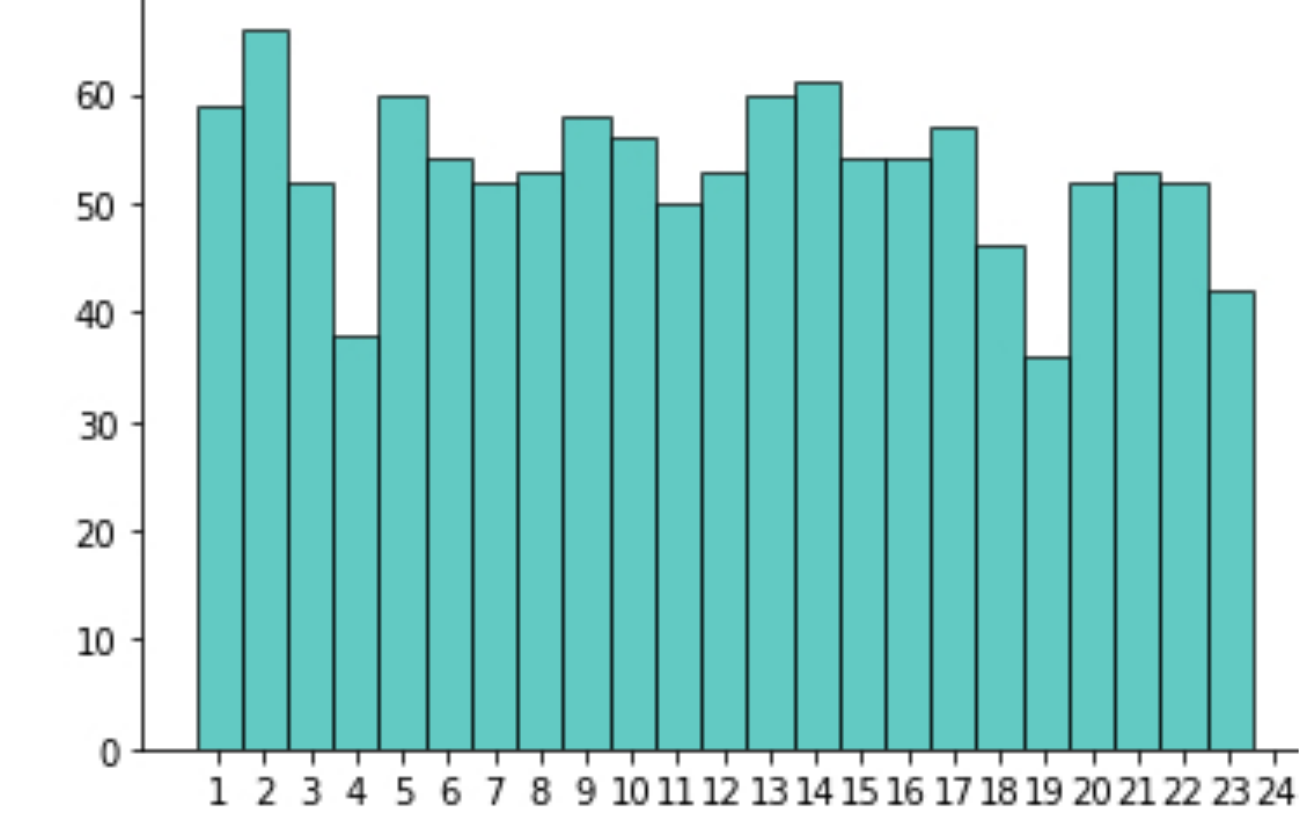}
        \caption{Percentage of overall shots}
    \end{subfigure}

    \caption{Percentage of different framing categories among photographs with people (the rest of the photographs are considered as medium shots)}
    \label{fig:range_ratios}
\end{figure*}

 \begin{figure}[htbp]
	\centering
	\includegraphics[width=\columnwidth]{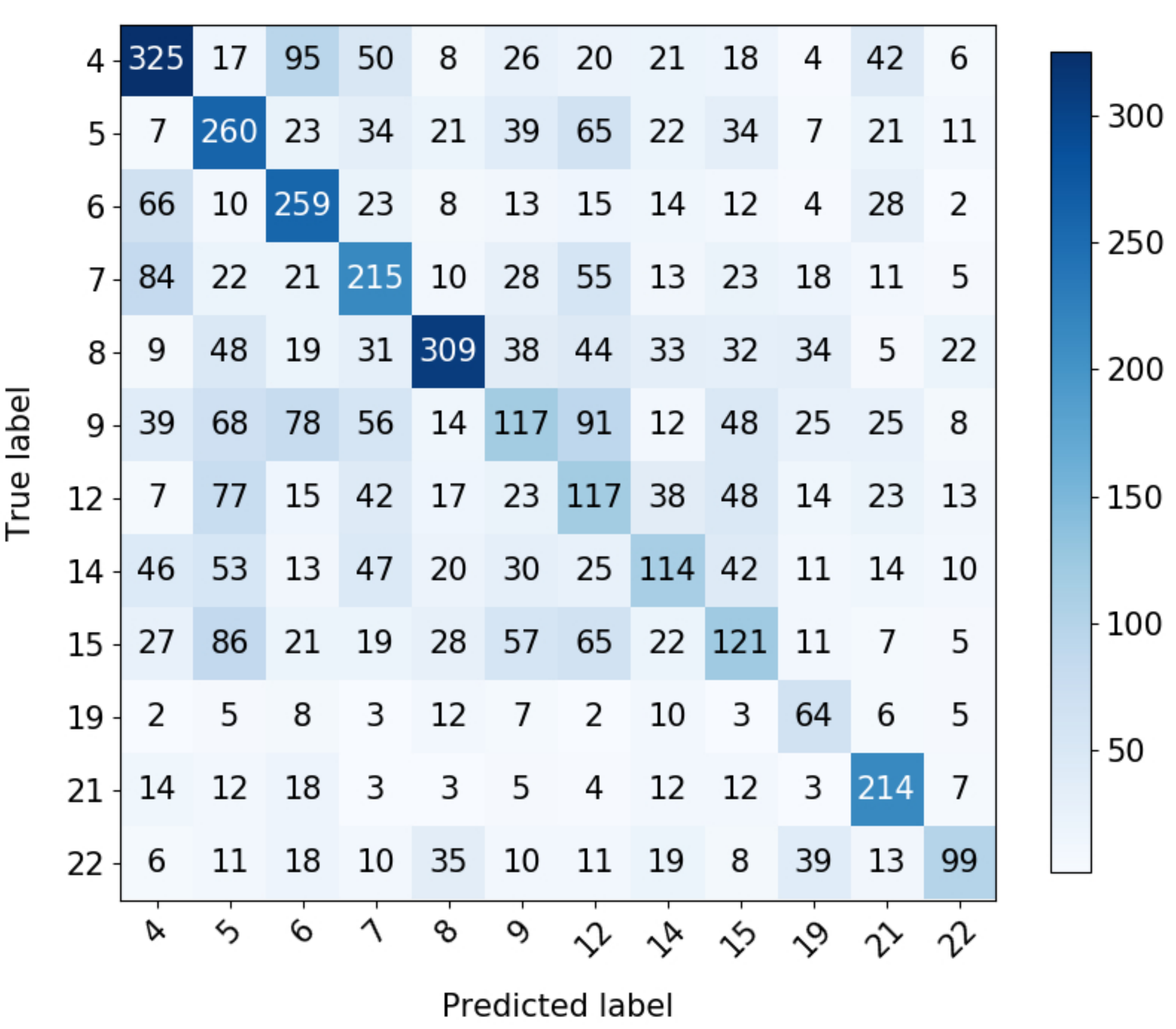}
	\caption{Confusion matrix for photographer recognition}
	\label{fig:confusion}
\end{figure}
\begin{figure*}[htbp]
	\centering
	\includegraphics[width=1.9\columnwidth]{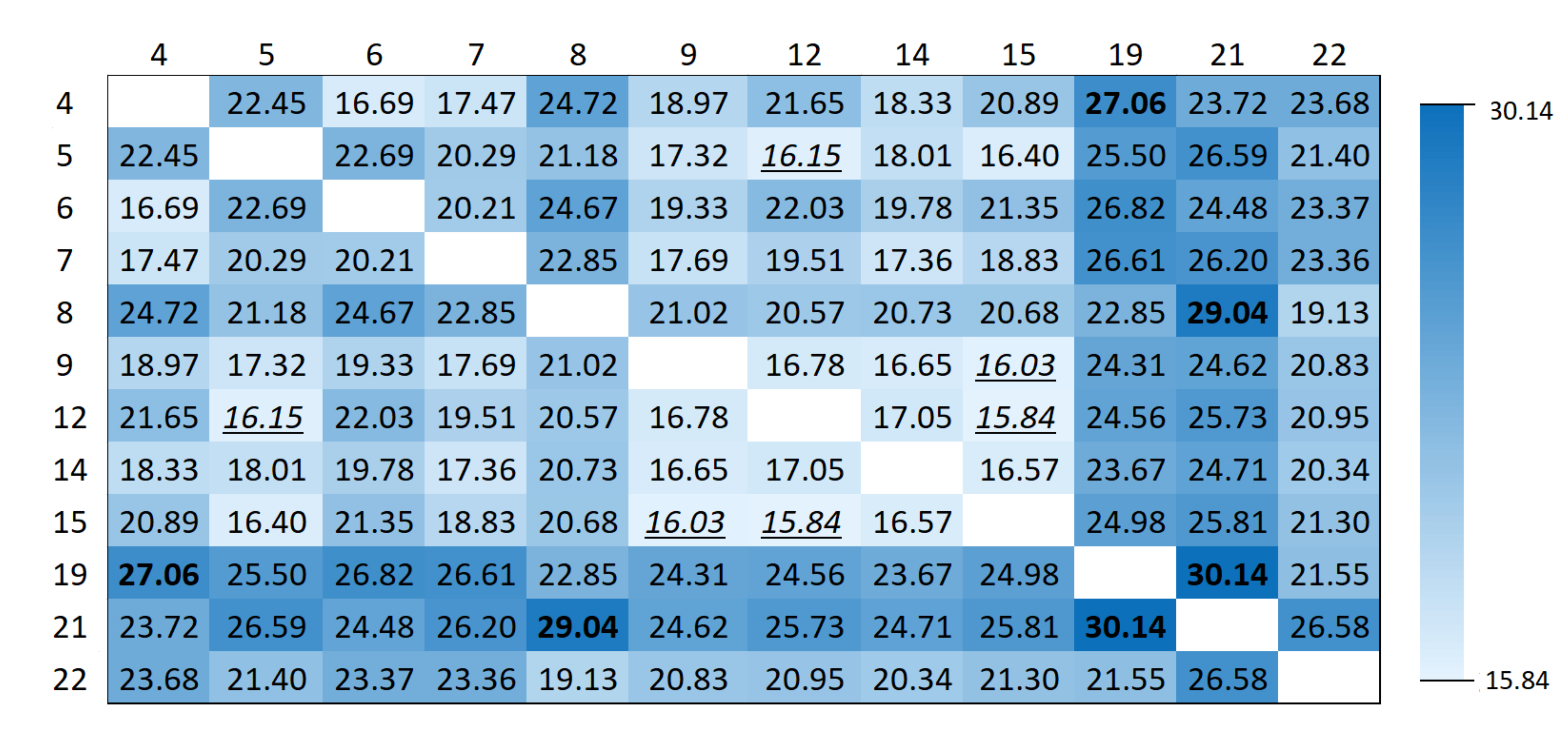}
	\caption{Similarity of the photographers (higher value denotes lower similarity).}
	\label{fig:wasserstein}
\end{figure*}

It is evident that the results do not provide exact object numbers. Instead, we exploit the results to evaluate relative numbers of occurrences of different objects in the photographs of each photographer. The object detection results for each photographer are given in Table~\ref{tab:detection}, where we report the ratio of images with people and the average number of persons in these images as well as the average number of occurrences of other objects per 100 images for each photographer and the average number of objects of all classes per 100 images.  For each object class, we highlight the values for photographers with the most frequent (bolded) and infrequent (italic) occurrences. We also provide the average frequency of each class among all photographers for reference.

As expected, we observe from Table~\ref{tab:detection} that different photographers concentrated on different content: 19-Sj\"oblom has people in 98\% of his images, while 10-Norjavirta and 14-Uomala have people in less than 85\% of their images. 8-Hedenstr\"om and 22-Manninen have the highest average number of people in these images (i.e., only images with people counted), while 6-Helander and 21-Roivainen captured images with fewer people. 

6-Helander and 15-Nurmi captured high numbers of airplanes, while 9-Suomela and 12-Kivi concentrated on boats. Interestingly, 6-Helander has rather high occurrence of other types of vehicles as well (boats, trains), while 15-Nurmi focused predominantly on airplanes. In 21-Roivainen's photos, there are many animals (horses, dogs), showing that many of his photos were taken in rural environements rather than urban scenes. 

Based on our manual inspection, chair pictures are typically taken indoors, while ties are worn by high ranking soldiers or wealthy people in urban conditions.Generally, presence of these objects in the scene allows us to make a conclusion that a photo is taken at some formal event. 19-Sj\"oblom, who has the highest ratio of photographs with people and 22-Manninen, who has the highest average number of people in his pictures, also have the most chairs. At the same time, both of them have a low rate of skis, dogs, and horses, supporting the claim that they were focusing on reporting the formal events and photographing high ranking military servants in urban environments. The connection between occurrence of chairs and people is supported by the fact that the  4-Hollming and 14-Uomala have the lowest chair rate. 14-Uomala has also a low ratio of people images, while 4-Hollming pictured a high amount of skiing photos, which shows that he photographed more outdoors. The occurrence of animals in his photos is rather high as well.

\subsection{Photo framing evaluation}
Photo framing evaluation is performed for photographs on which people are present according to the output of the combination of object detectors. We manually defined two thresholds to divide such photographs into three classes: close-ups, medium shots, and overall shots. Fig.~\ref{fig:ranges} shows an example photograph belonging to each of these classes.  

Fig.~\ref{fig:range_ratios} shows how the photographs with people are divided into different framing categories for different photographers (the percentages of close-ups and overall shots are shown, the remaining percentage corresponds to medium shots). The figure shows that 19-Sj\"oblom took relatively most close-ups and medium shots and fewest overall shots. From the previous subsection, we know that he had also the highest ratio of photos with people and the objects detected in his photographs profiled him as an urban photographer. 

We also observe that the rest of the three photographers having the highest ratio of close-ups, i.e., 6-Helander, 7-J\"anis, and 21-Roivainen have also rather low chair and tie rates along with a low number of persons per image. At the same time the number of images on which people are present is average for these photographs. These observations lead us to the conclusion that these photographers were focusing on portrait photographs when photos of people were taken.

In addition, we observe that 2-Nousiainen and 14-Uomala captured relatively most overall shots. 14-Uomala also had fewest people photographs in general and only few chairs in his images, which led us to conclude that he did mostly outdoor photography. The fact that the average number of objects on his photographs is also the lowest leads us to the same conclusion. These observations support each other as overall shots are mainly outdoor images. 18-Laukka took fewest close-ups, and the objects detected in his photos mainly profile him as a non-urban photographer, although the ratio of objects in each category is rather low. Interestingly, 4-Hollming had only few overall shots, while the object in his photographs profiled him as a non-urban outdoor photographer. One possible explanation comes from the fact that he has the highest ratio of skis, which are more likely to be present in close-up or medium photographs of people that are taken outdoors.
\begin{table}[tb]
\caption{Classification accuracies of different photographers}
\centering
\begin{tabular}{l|c}
Photographer ID & Accuracy \\
\hline
4-V\"ain\"o Hollming     & 51.4\% \\
5-Jarl Taube             & 47.8\%  \\
6-Nils Helander          & 57.1\% \\
7-Pauli J\"anis             & 42.6\%  \\
8-Oswald Hedenstr\"om        & 49.5\% \\
9-Esko Suomela           & 20.1\% \\
12-Kauko Kivi             & 26.9\% \\
14-Vilho Uomala           & 26.8\%  \\
15-Eino Nurmi            & 25.8\%  \\
19-Kalle Sj\"oblom          & 50.4\% \\
21-Heikki Roivainen        & 69.7\%  \\
22-Esko Manninen        & 35.5\%  \\
\end{tabular}
\end{table}

\subsection{Photographer recognition}
Following the described classification method for photographer recognition, we evaluated whether the trained network can be used to recognize the photographer for the unseen photographs not used in training. Since each photographer has a certain amount of duplicate images, here we split the photographs into train and test sets according to the capturing times to ensure that photographs depicting the same event are not used for both training and testing.

Overall, the network achieved 41.1\% classification accuracy on the test set.  The confusion matrix of the classification results in shown in Fig.~\ref{fig:confusion}, where all the diagonal elements represent correctly classified samples. We see that the network was able to correctly classify a significant part of photographs from each of the photographers. The photographer-specific recognition rates vary from 20.1\% for 9-Suomela to 69.7\% for 21-Roivainen. The recognition accuracy of each photographer is shown in Table 3.

Comparison of the recognition results with the earlier analysis on detected object reveals that some of the most recognized photographers have also specific objects. 21-Roivainen (69.7\% accuracy) has most dogs, horses, and cars in his pictures. 4-Hollming (51.4\%) has the highest number of skiing pictures and only few chairs (i.e., many outdoor photos). 22-Manninen (35.5\%) had the highest average number of people in his people photos and the highest occurrence of chairs (i.e., indoor photos). 19-Sj\"oblom (50.4\%) captured photographs in urban environments. Some of the main confusions occur between 4-Hollming, 6-Helander, and 7-J\"anis. We observe also that all three of them can be considered as non-urban photographers having rather similar number of persons per image and relatively low number of chairs and ties.

In addition, 5-Taube and 12-Kivi are confused to each other. 19-Sj\"oblom and 22-Manninen are often misclassified as 8-Hedenstr\"om - these are also the three photographers having the highest numbers of chairs and ties, i.e., the photographers that appear to be the most urban. Also 9-Suomela is often misclassified as 12-Kivi, - both of them have a high rate of boats and rather high rate of trains. These observations support the conclusion that besides establishing the authorship of photographs, the learned feature representation allows to make conclusions about overall visual similarity of these photographers and similarity of the styles of their photos.

 \begin{figure*}[htbp]
	\centering
	\includegraphics[width=1.9\columnwidth]{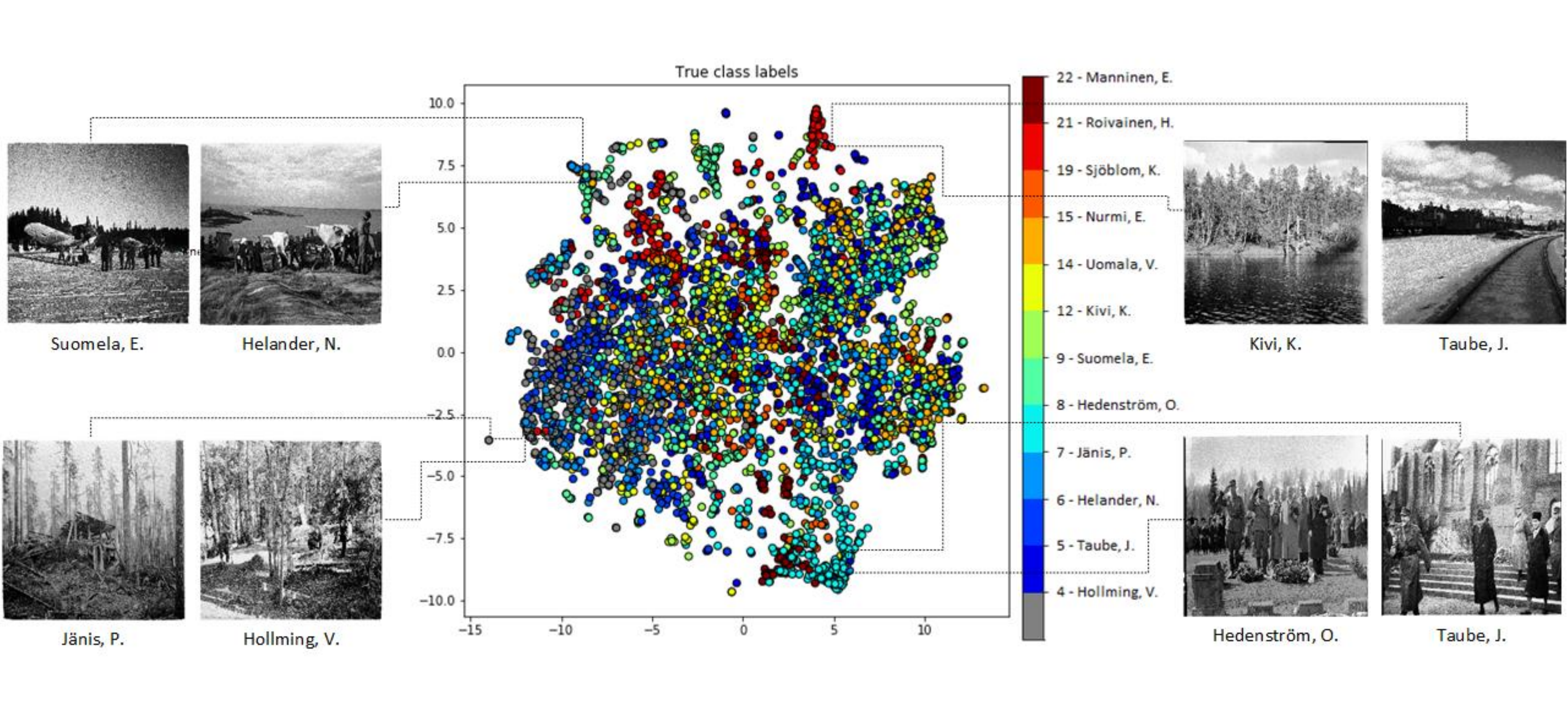}
	\caption{Visualization of the photograph similarities using the t-SNE algorithm and sample photographs with a varying similarity. Source of photographs: SA-kuva.}
	\label{fig:tSNE}
\end{figure*}
\subsection{Photographer visual similarity}
We calculated the Earth Mover's Distance between all pairs of photographers to assess their similarity as described in section III-C. The results are outlined in Fig.~\ref{fig:wasserstein}, where higher values show higher distances, i.e., lower similarity. The highest distances are highlighted in bold, and the lowest distances are underlined. We observe that the obtained results correspond closely to the misclassification rates of the photographers, i.e., photographers that are often misclassified as each other also have low distance between each other. For example, this can be observed in the pairs 5-Taube and 12-Kivi; 15-Nurmi and 9-Suomela; 12-Kivi and 15-Nurmi. We also observe that for these photographers the similarities can be seen based on the detected objects as well: both 15-Nurmi and 12-Kivi have similar number of objects per image and person images, and their rates of persons per image are equal. We observe the same for the pair 15-Nurmi and 9-Suomela.

Besides, observing the photographers that are identified to be distant from each other, we can see that their mutual misclassification rates are low, and the objects detected on their photographs provide a reasonable explanation to their differences. The three least similar pairs are 19-Sj\"oblom and 21-Roivainen; 21-Roivainen and 8-Hedenstr\"om; 19-Sj\"oblom and 4-Hollming. From Table 2, we can observe that 19-Sj\"oblom has the highest ratio of person images, the lowest number of dogs and horses, and the highest number of chairs and ties. At the same time, 21-Roivainen has the highest ratio of dogs and horses, and low numbers of chairs and ties. He also has an average number of person images and objects per image, while 19-Sj\"oblom has rather high ratios. The similar reasoning holds for the pair 4-Hollming and 19-Sj\"oblom, as 4-Hollming has low number of chairs and ties, while having rather high ratios of non-urban photos. Besides, 4-Hollming has the highest ratio of skiing photos, meaning that he was most likely having significantly more winter photos than other photographers, making him visually more distinguishable. 21-Roivainen has the lowest ratio of persons per image, while 8-Hedenstr\"om has the highest. The average number of objects per image differs rather significantly as well. Besides, 8-Hedenstr\"om has high ratios of chairs and ties compared to 21-Roivainen.

Another noticeable fact is that the photographers having many of the extreme values of detected objects, e.g., 19-Sj\"oblom (maximal ratio of person images, chairs, and ties, and minimal ratios of dogs and horses) and 21-Roivainen (maximal ratios of cars, dogs, and horses; minimal ratios of persons per image and bicycles) also have the higher distances to all other photographers, which can be seen by the fact that the corresponding rows and columns are rather dark compared to others. These facts support the meaningfulness of the proposed method for establishing photographers' similarity.

We further examined the similarities and differences between the photographers by extracting the features learned by the classifier network for the test images and visualize them using the t-SNE algorithm \cite{tSNE} in Fig.~\ref{fig:tSNE}. In the figure, the dots denote photographs and different colors correspond to different photographers. Some of the colors are clearly concentrated on certain spots further confirming that different images are characteristic for different photographers. The similarities of adjacent images are also illustrated by the examples, showing that images that are located close to each other in the feature space are also visually similar even if the photographers are different. It can be observed that the upper-right corner images represent the landscapes, the lower-right corner shows images that contain many people in close proximity, and the upper-left corner contains images of multiple people located at distance, where not much other details are present. This further confirms the reasonability of the proposed approach for obtaining feature representations and allows to compare visual similarity of images of different photographers.

\section{Conclusion}
\label{sec:conclusion}

We showed that modern machine learning algorithms can help in societal research on historical photo archives in many ways. In this paper, we applied state-of-the-art object detection models and neural network architectures to obtain statistics and characteristics of prominent Finnish World War II photographers. We examined the typical object categories in the photos of each photographer and analyzed the differences in their ways of capturing and framing people. Furthermore, we showed that a convolutional neural network was able to some extent recognize photographers from the photos leading to the conclusion that certain photos can be considered typical for a specific photographer. The confusion matrix of the photographer classifier revealed some similarities between the photographers. We are not aware of any prior works using machine learning to such photographer analysis. The obtained results will help historians, other researchers, and professionals using historical photo archives in their work to analyze and compare the works of specific photographers.

In this work, we used only publicly available pretrained object detection models and basic photograph information, i.e., the photographer, available for most photo archives. The pretrained models showed good performance on the historical gray-scale photographs even though pretrained with modern color photos. We also provide all codes, models and data annotations along with a detailed description on how to use them. Thus, the same methods can be easily applied on other historical photo archives.  At the same time, we remind that, in a single paper, we can demonstrate only a tiny fraction of all currently available machine learning tools. The photographer analysis could be further enhanced, for example, by considering photographer intentions \cite{intention} and their photo quality \cite{photoquality}, besides, object detection performance can be enhanced by considering information fusion approaches \cite{chen2019deep}, as well as improving detection of smaller sized objects \cite{cao2019improved}. Here, we only considered person detection, while person segmentation \cite{jiang2019end}, face detection with facial expression analysis \cite{uddin2017facial}, group-level emotion recognition \cite{yu2019group}, or age estimation \cite{al2019comprehensive} will open more interesting opportunities. Besides object-level analysis, scene recognition \cite{zheng2017multicriteria} will help to further characterize photographers.  

In the future, we will concentrate on issues requiring more specialized methods such as recognizing object classes only appearing in Finnish historical photos or during World War II. We aim at exploiting the original textual photo descriptions to produce more complete object labeling and as well as topic and event recognition \cite{zhang2017automatic}. This will help us to solve one of the biggest challenges in analysing wartime photos, namely separating different statuses of subjects - whether the people in the photographs are alive, wounded or deceased. These kinds of more refined results can help us in the end to draw a more detailed picture of the aims, qualities, and characters of individual TK photographers. We aim at publishing all our results in the archive to assist different types of societal studies on the archive. 
\bibliographystyle{plain}
\bibliography{sample}
\clearpage
\begin{IEEEbiography}[{\includegraphics[width=1in,height=1.25in,clip,keepaspectratio]{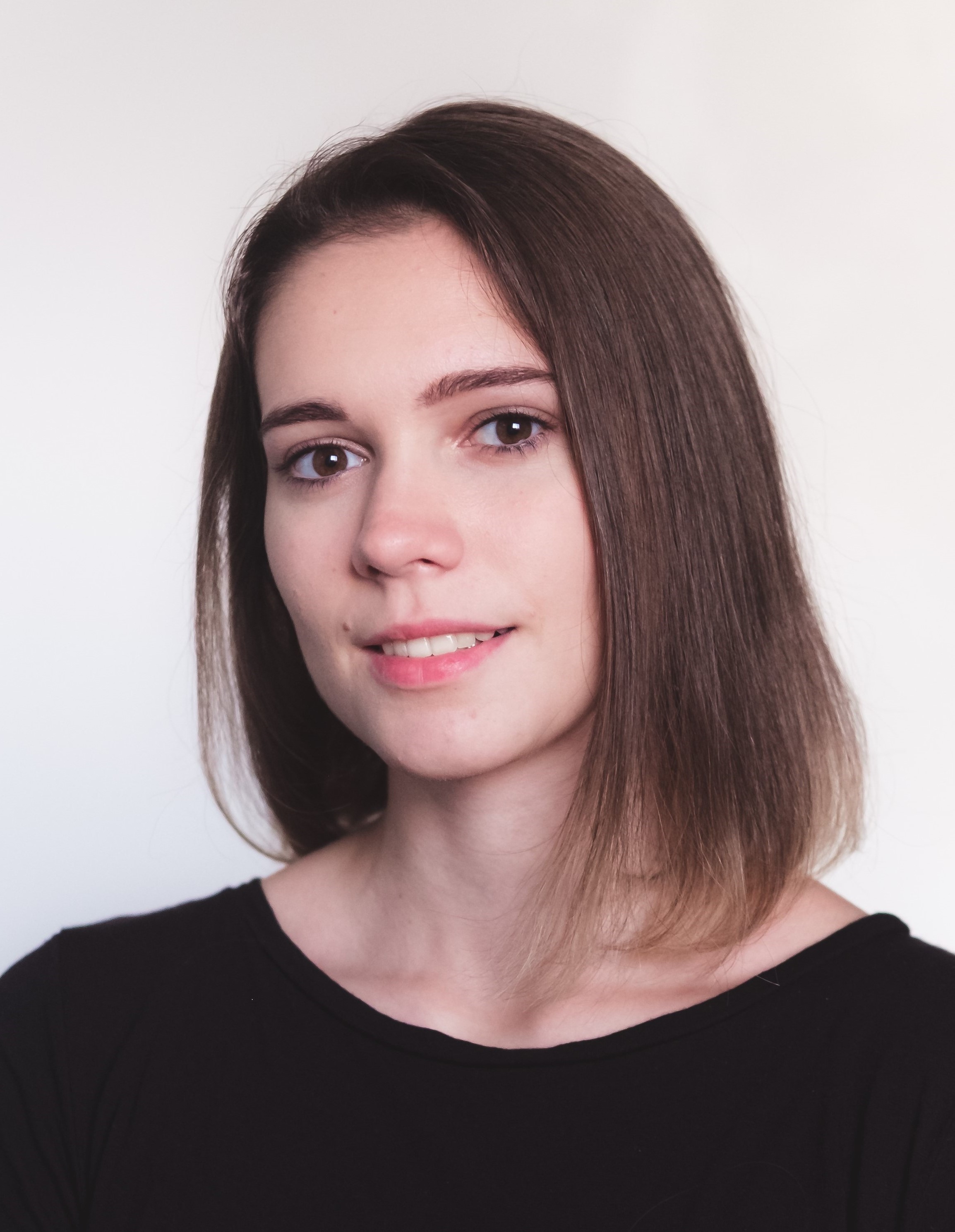}}]{Kateryna Chumachenko} received her MSc. degree from Tampere University, Finland in 2019 majoring in Data Engineering and Machine Learning, and BEng. degree from South-Eastern Finland University of Applied Sciences, Mikkeli, Finland in 2017. She is currently a doctoral student within Signal Analytics and Machine Intelligence research group at the Faculty of Information Technology and Communication Sciences in Tampere University, Finland. Her research interests lie in the fields of machine learning and pattern recognition including statistical learning methods and applications in computer vision.
\end{IEEEbiography}

\begin{IEEEbiography}[{\includegraphics[width=1in,height=1.25in,clip,keepaspectratio]{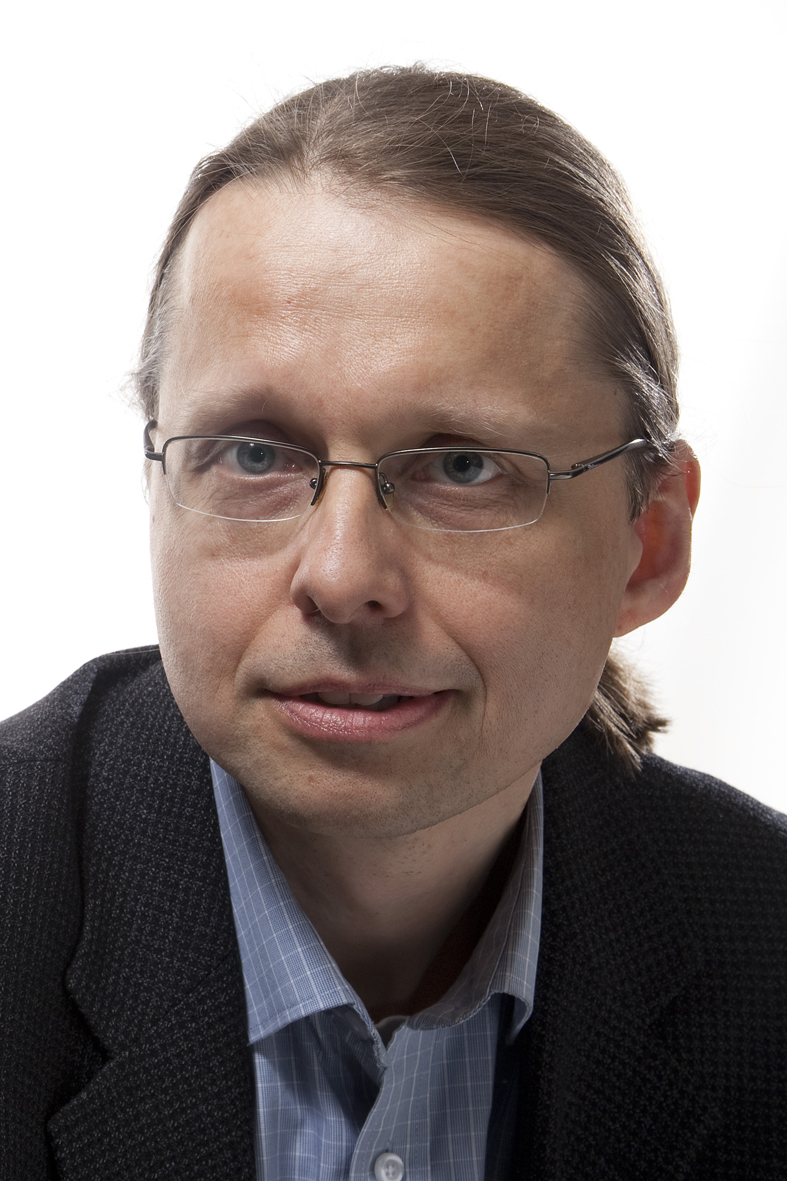}}]{Anssi M\"annist\"o} received his Ph.D. degree in the Department of Social Sciences at the University of Tampere, Finland in 2000. Since 2003 he has worked as a senior lecturer of visual journalism, currently in the Faculty of Information Technology and Communication
Sciences, Tampere University, Finland. His research interests include new modes of journalism and photojournalism, use of modern data visualizations in journalism and education and developing methods in quantitative and content analyze of photographs.
 \end{IEEEbiography}

\begin{IEEEbiography}[{\includegraphics[width=1in,height=1.25in,clip,keepaspectratio]{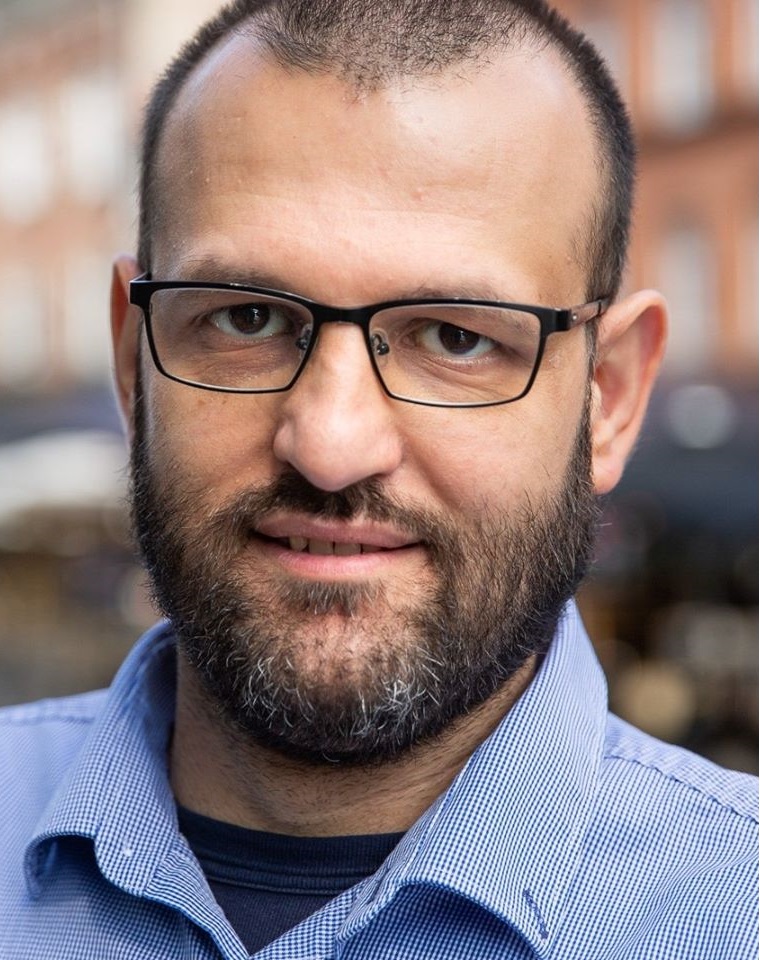}}]{Alexandros Iosifidis} (SM'16) is leading the Machine Learning and Computational Intelligence group at the Department of Engineering, Aarhus University, Denmark. He received the B.Sc. degree in Electrical and Computer Engineering and the M.Sc. degree with a specialisation in Mechatronics from the Democritus University of Thrace, Greece, in 2008 and 2010, respectively. He also received a PhD in Computer Science from the Aristotle University of Thessaloniki, Greece, in 2014. Before he joins Aarhus University, he held Postdoctoral Researcher positions at Tampere University of Technology, Finland, where he was an Academy of Finland Postdoctoral Research Fellow.

Dr. Iosifidis has contributed in more than twenty R\&D projects financed by EU, Finnish and Danish funding agencies and companies. He has (co-)authored 68 articles in international journals and 91 papers and abstracts in international conferences proposing novel Machine Learning techniques and their application in a variety of problems. He is a Senior Member of IEEE since 2016, and he served as an Officer of the Finnish IEEE Signal Processing-Circuits and Systems Chapter during 2016-2018. He is currently a member of the EURASIP Technical Area Committee on Visual Information Processing, and serves as Area/Associate/Academic Editor for Neurocomputing, Signal Processing: Image Communications, IEEE Access and BMC Bioinformatics journals. His research interests focus on topics of neural networks and statistical machine learning finding applications in computer vision, financial modeling and graph analysis problems.
\end{IEEEbiography}

\begin{IEEEbiography}[{\includegraphics[width=1in,height=1.25in,clip,keepaspectratio]{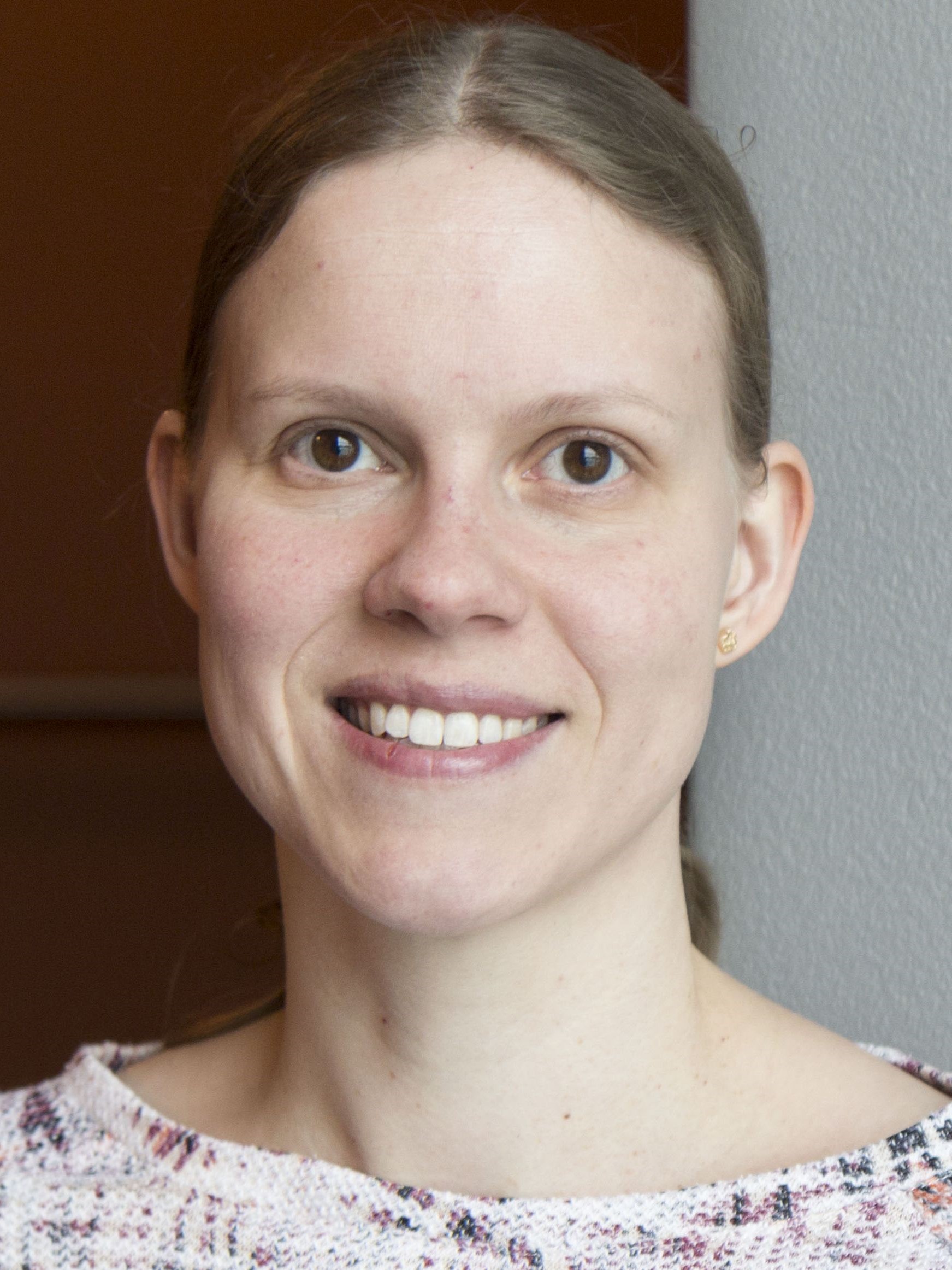}}]{Jenni Raitoharju} received her Ph.D. degree at Tampere University of Technology, Finland in 2017. Since then, she has worked as a Postdoctoral Research Fellow at the Faculty of Information Technology and Communication Sciences, Tampere University, Finland. In 2019, she started working as a Senior Research Scientist at the Finnish Environment Institute, Jyväskylä, Finland after receiving Academy of Finland Postdoctoral Researcher funding for 2019-2022. She has co-authored 13 journal papers and 27 papers in international conferences. She is the chair of Young Academy Finland 2019-2020. Her research interests include machine learning and pattern recognition methods along with applications in biomonitoring and autonomous systems.
\end{IEEEbiography}

\EOD
\end{document}